\documentclass[10pt,twocolumn,letterpaper]{article}

\usepackage{cvpr}              

\usepackage[table]{xcolor}
\usepackage{multirow}
\usepackage{booktabs}
\usepackage{makecell}
\usepackage{amssymb}      
\usepackage{pifont}       
\usepackage{siunitx}
\newcommand{\xmark}{\ding{55}}

\newcommand{\name}{\textit{CrowdEraser}} 
\newcommand{\dsname}{\textit{EgoCrowds}} 

\newcommand{\best}[1]{\cellcolor{red!16}\textbf{#1}}
\newcommand{\second}[1]{\cellcolor{yellow!25}#1}

\usepackage{colortbl,xcolor}
\definecolor{bestcolor}{RGB}{198,239,206}      
\definecolor{secondcolor}{RGB}{255,235,156}    

\definecolor{cvprblue}{rgb}{0.21,0.49,0.74}
\usepackage[pagebackref,breaklinks,colorlinks,allcolors=cvprblue]{hyperref}

\def\paperID{****} 
\def\confName{CVPR}
\def\confYear{2026}

\title{Generating Humanless Environment Walkthroughs from \\ Egocentric Walking Tour Videos}

\author{Yujin Ham \quad Junho Kim  \quad  Vivek Boominathan  \quad  Guha Balakrishnan \vspace{1.5mm} \\
Rice University \vspace{1.5mm} \\
{\tt\small \{yh106, jk84, vivekb, guha\}@rice.edu}
}

\begin{document}

\twocolumn[{
\renewcommand\twocolumn[1][]{#1}%
\maketitle
\vspace{-0.2in}
\begin{center}
\vspace{-7pt}
    \centering \includegraphics[width=\textwidth]{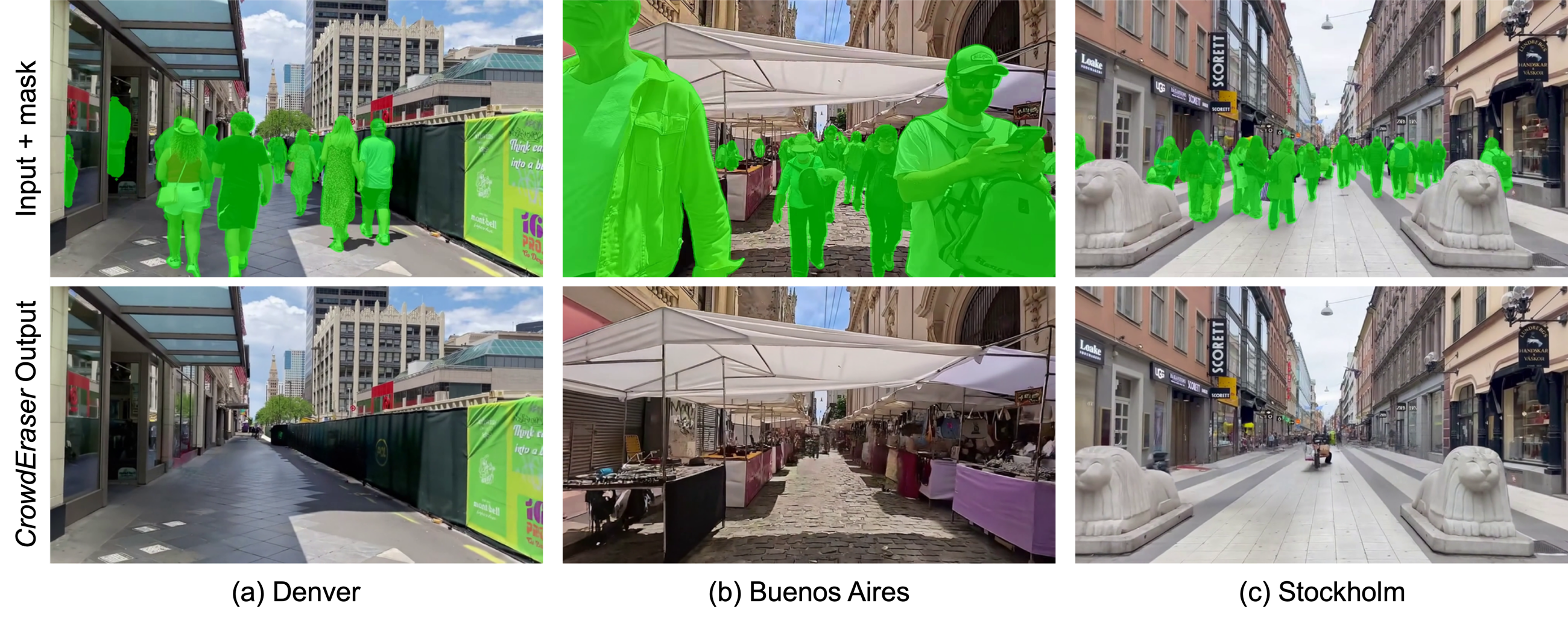}
    \vspace{-0.2in}
    \captionof{figure}{\textbf{Example results of \name{}, the proposed method in this study, for removing humans and their associated effects from three egocentric ``walking tour'' video clips.} The model takes a video clip along with foreground human masks as input (top). \name{} generates a new video clip with the humans and their shadows removed. \name{} works well when confronted with significant human presence due to (a,c) crowds, and (b) proximity of others to the camera wearer. Comparisons with baseline methods for these scenes are provided in the supplementary material.}
    \label{fig:teaser}
\end{center}%
}]

\begin{abstract}
Egocentric ``walking tour'' videos provide a rich source of image data to develop rich and diverse visual models of environments around the world. However, the significant presence of humans in frames of these videos due to crowds and eye-level camera perspectives mitigates their usefulness in environment modeling applications. We focus on addressing this challenge by developing a generative algorithm that can realistically remove (i.e., inpaint) humans and their associated shadow effects from walking tour videos. Key to our approach is the construction of a rich semi-synthetic dataset of video clip pairs to train this generative model. Each pair in the dataset consists of an environment-only background clip, and a composite clip of walking humans with simulated shadows overlaid on the background. We randomly sourced both foreground and background components from real egocentric walking tour videos around the world to maintain visual diversity. We then used this dataset to fine-tune the state-of-the-art Casper video diffusion model for object and effects inpainting, and demonstrate that the resulting model performs far better than Casper both qualitatively and quantitatively at removing humans from walking tour clips with significant human presence and complex backgrounds. Finally, we show that the resulting generated clips can be used to build successful 3D/4D models of urban locations. More results and code are available at \texttt{https://crowd-eraser.github.io/}
\end{abstract}


\section{Introduction}
\label{sec:intro}
High-fidelity models of everyday urban environments, from city streets to building lobbies, are essential for driving progress across various computer vision domains, including 3D neural rendering, robotics, user content generation, and autonomous driving. Perhaps the richest and most widely available sources of diverse urban environment imagery are ``walking tour'' videos shared on platforms such as YouTube, which depict the experience of walking through an environment from an egocentric (i.e., first person) viewpoint. Thousands of hours of walking tour footage exist that span most countries around the world. However, these videos have a critical drawback that prevents their direct use in static environment extraction: the significant presence of human transients occupying a large fraction of pixels per frame and occluding the scene structure. Typical urban scenes feature large pedestrian groups that collectively occlude many pixels (see Fig.~\ref{fig:teaser}a), and due to the ground-level egocentric perspective, even a single individual traversing close to the camera wearer can monopolize a substantial pixel area (see Fig.~\ref{fig:teaser}b). 

In this study, we address this challenge by developing an algorithm (\name{}) that can realistically remove humans and their associated effects (e.g., shadows, accessories) from egocentric walking tour videos (see Fig.~\ref{fig:teaser}). We empirically found that Casper, the powerful object-effect-removal diffusion model developed for GenOmnimatte~\cite{Lee_2025_CVPR}, performs reasonably well on videos depicting few humans at a sufficient distance from the camera. However, Casper produces unreasonable artifacts when faced with significant human presence and complex outdoor backgrounds. Hypothesizing that this performance gap is mainly caused by a domain shift between Casper's training dataset and walking tour data rather than a shortcoming of the diffusion model design itself, we focused this project on the careful development of a rich supervised training dataset for person (and shadow) inpainting in walking tour videos. 

Our main contribution is the construction of \dsname{}, a dataset consisting of 1,000 pairs of diverse 7-second walking tour video clips with and without humans and their associated shadows. Constructing \dsname{} with real video pairs would require highly controlled setups with human actors and camera rigs (to exactly duplicate camera trajectories), which is neither feasible to capture nor likely to result in sufficient visual diversity. Instead, inspired by the successes of semi-synthetic training datasets for other tasks such as optical flow estimation~\cite{dosovitskiy2015flownet} and image segmentation~\cite{zhao2019data}, we devised a semi-synthetic video generation pipeline to construct \dsname{}. We first curated a large corpus of 7-second clips from walking tour videos spanning 50 cities around the world (see Fig.~\ref{fig:location-map}). Using automated recognition and segmentation algorithms~\cite{ren2024grounded, liu2024grounding, ravi2024sam}, we separated these clips into ``empty'' background clips consisting of few-to-no humans, and ``crowd'' clips with significant human presence, and synthesized 1,000 composited videos of random human foregrounds overlaid on backgrounds. In addition, we added per-human associated shadow effects using simple rule-based procedures with random augmentations. This dataset synthesis approach allows us to generate videos retaining the appearance, occlusion, and motion realism (including camera motion) of naturally captured footage while providing the necessary ground-truth supervision for model training. 
We finetuned Casper on pairs of these composited videos and their corresponding ``empty'' versions (background only) in \dsname{}, resulting in an inpainting model (which we call \name{}) optimized for walking tour videos. 

We evaluated \name{} qualitatively and quantitatively on the diverse \dsname{} test set. We quantitatively compared \name{} to competing inpainting methods for human removal: GenOmnimatte~\cite{Lee_2025_CVPR}, ProPainter~\cite{zhou2023propainter}, and DiffuEraser~\cite{li2025diffueraser}. Results demonstrate that \name{} offers a clear improvement over these baselines in terms of standard reconstruction metrics (PSNR, LPIPS), particularly when the percentage of frame pixels occluded by humans (which we denote ``\emph{Crowd\%}'') increases. We further present qualitative results on real walking tour videos. Finally, we demonstrate that the humanless outputs generated by \name{} enable realistic 3D environment modeling using a state-of-the-art 3D reconstruction method~\cite{xiao2025spatialtrackerv2}, which was not possible using the original videos alone.

\section{Related Work}
\label{sec:related}

\subsection{Video Matting and OmniMatte}
There is a rich history in computer vision and graphics on representing videos as a set of component layers~\cite{brostow1999motion,fradet2008semi,gandelsman2019double,jojic2001learning,pawan2008learning,wang1994representing}. Video matting works decompose a video into foreground and background layers that may be alpha blended together to recover the original videos~\cite{bai2009video, chuang2002video, hou2019context,li2005video,sengupta2020background,wang2005interactive,xu2017deep}. Of these, the recent Omnimatte video matting line of studies is most relevant to our work~\cite{lin2023omnimatterf,Lee_2025_CVPR,Lu_2021_CVPR,Suhail_2023_CVPR}. The goal of Omnimatte methods is to decompose videos into individual object layers along with their associated scene effects (e.g., shadows, accessories, reflections), with the background being a key layer independent of all foreground elements. The original Omnimatte uses optical flow-based optimization, assumes a static background and approximates camera motion via a homography mapping each video frame to a canonical \textit{canvas} image~\cite{Lu_2021_CVPR}. Omnimatte3D~\cite{Suhail_2023_CVPR} relaxes the planar-homography assumption by predicting per-frame background and disparity maps, using multi-view consistency to handle parallax and non-planar scenes. OmnimatteRF~\cite{lin2023omnimatterf} further models the static background in 3D using a radiance field trained while masking out foreground regions, relieving the camera motion conditions. 
Finally, the most recent Generative Omnimatte~\cite{Lee_2025_CVPR} employs a powerful video diffusion backbone~\cite{bar2024lumiere, ma2024vidpanos} trained on a curated dataset to perform video matting using strong generative and semantic priors. While the most powerful of all Omnimatte methods, Generative Omnimatte still struggles in recovering clean backgrounds on videos with significant human-masked regions.

\subsection{Video Inpainting}
Video inpainting methods aim to fill in masked spatiotemporal regions in videos while maintaining coherence with the visible scene context. Classical (pre-deep learning) approaches leverage techniques such as optical flow estimation~\cite{matsushita2006full,patwardhan2005video}, energy-based optimizations~\cite{bertalmio2001navier,ebdelli2015video}, and exemplar-based strategies~\cite{ghanbari2011contour,newson2014video,shih2009exemplar}. However, these methods are limited in their modeling capacity and often produce unrealistic results. Current state-of-the-art video inpainters all rely on deep generative models. Propainter~\cite{zhou2023propainter} is a transformer model that is optimized directly for video inpainting. Several other inpainting methods~\cite{bian2025videopainter, ding2025homogen, Lee_2025_CVPR, lee2025video, li2025diffueraser, liu2025generative, ma2024vidpanos} leverage ``foundation'' generative models such as Stable Diffusion~\cite{rombach2022high, zhang2023adding, podell2023sdxl} for images and Lumiere~\cite{bar2024lumiere} for videos. We build on Casper, the diffusion model developed in one of these studies~\cite{liu2025generative} for our work, by extending its capabilities for removing people and associated effects from walking tour videos.
\subsection{Egocentric Walking Videos}
Walking tour videos are a relatively untapped data source in the computer vision community, with the exception of a few recent studies~\cite{li2025sekai,VenkataramananR24} and broader egocentric datasets~\cite{grauman2022ego4d}. These studies focused on using these videos for visual recognition tasks such as action recognition, object interaction, and scene understanding, but have not tackled scene inpainting. A related video dataset is MannequinChallenge~\cite{li2019learning}, which consists of videos captured from a moving first-person handheld camera and depicting scenes of people performing the once-viral trend of ``freezing'' in place. These videos are uniquely valuable for learning human-scene decompositions, but are limited in content to the time period and locations of this social trend. 


\section{Construction of \dsname{} Dataset}
\label{sec:data}
\begin{figure}[t!]
    \centering
    \includegraphics[width=0.9\linewidth]{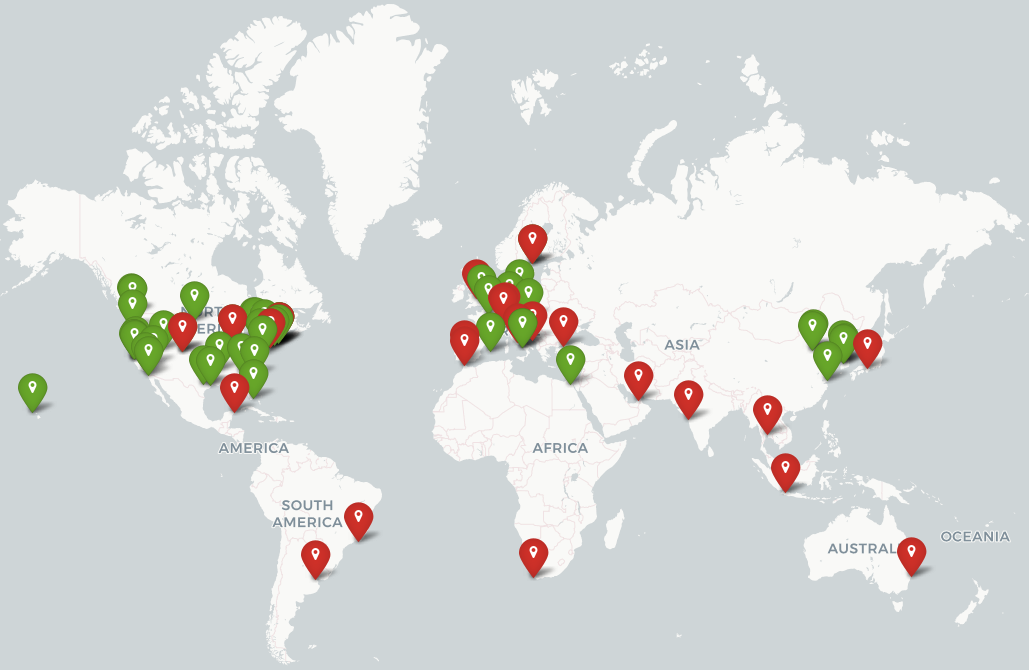}
    \caption{\textbf{Locations of background video clips in \dsname{}.} Training clip locations are in green, and testing locations are in red. The full list of city names are in Supplementary.}
    \label{fig:location-map}
\end{figure}
\begin{figure*}[t!]
    \centering
    \includegraphics[width=\textwidth]{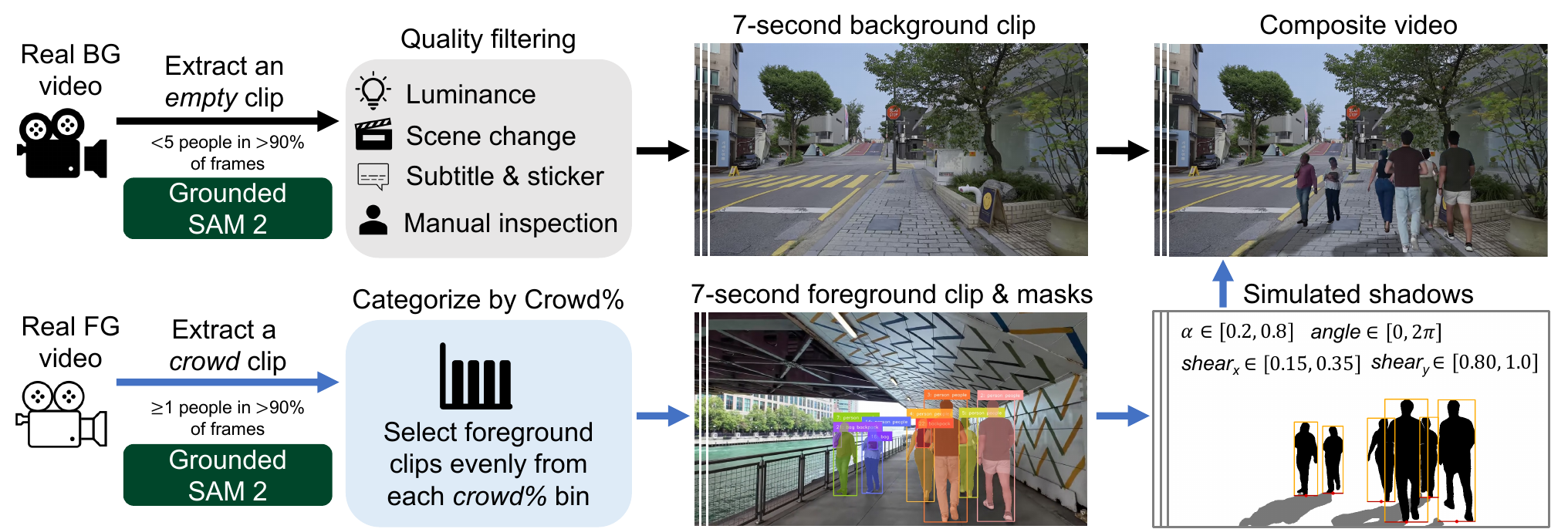}
    \caption{\textbf{Overview of our data construction pipeline.} Both background and foreground clips are sourced from real ``\textit{walking tour}'' videos. The foreground clips were selected to ensure an approximately uniform distribution across different \emph{Crowd\%} levels. For each instance, we generate a soft shadow with randomized strength and angle by applying an affine transform to the human mask (red dots indicate pivot points).}
    \label{fig:data_pipeline}
\end{figure*}
\begin{figure}[t!]
    \centering
    \includegraphics[width=\linewidth]{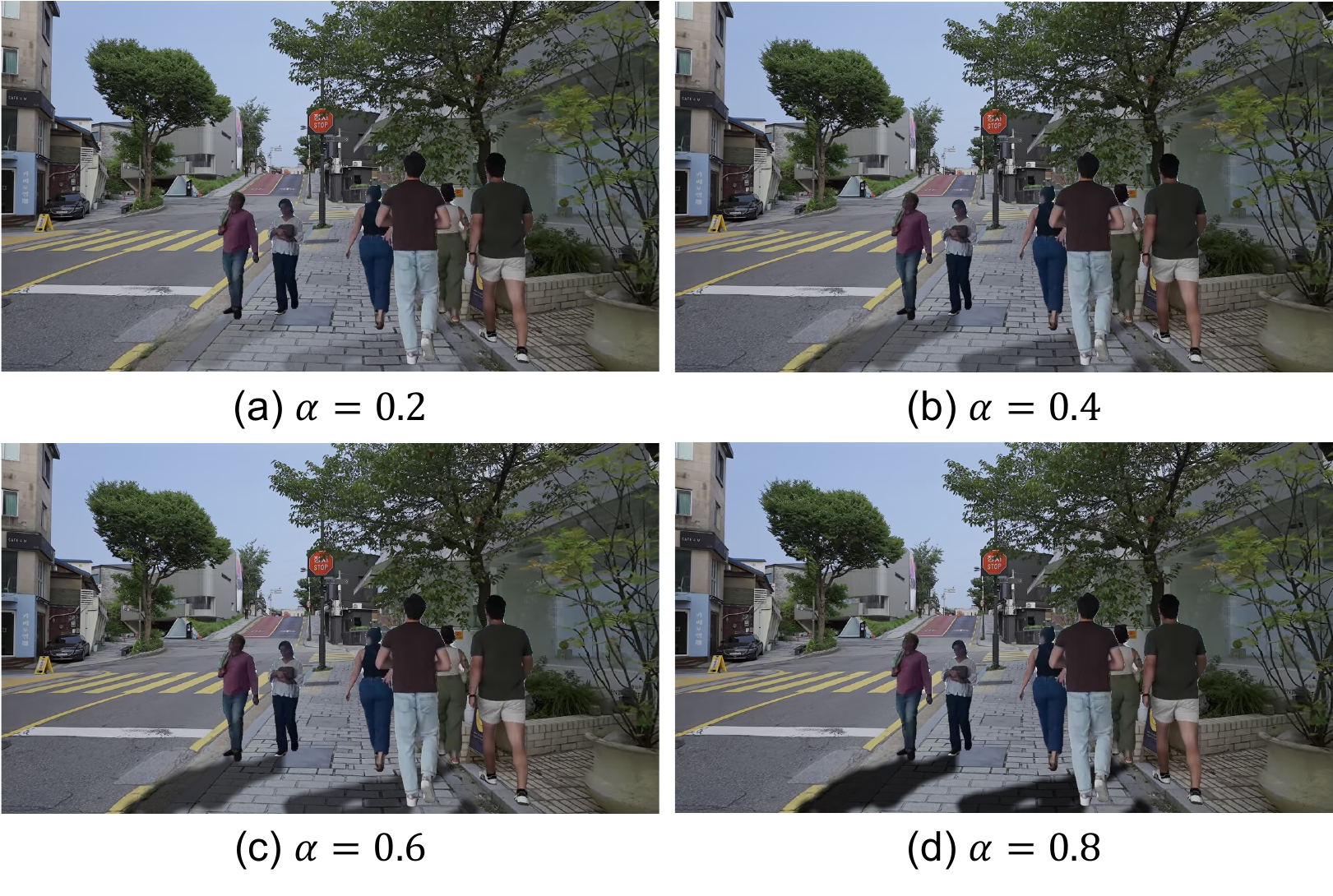}
    \vspace{-0.25in}
    \caption{\textbf{Shadow injection with varying $\alpha$ values.} As $\alpha$ ranges from 0.2 (a) to 0.8 (d), the shadow intensity appears stronger.}
    \label{fig:shadow-alpha}
\end{figure}


There is no existing dataset of real video clip pairs depicting a scene with and without human crowds, particularly from a first-person perspective. Because it is practically impossible to capture the same scene with and without humans under identical lighting and camera motion conditions, we constructed \dsname{} in a semi-synthetic manner by carefully compositing background and foreground components of real walking tour video clips sourced from the web. Each video clip is 7 seconds (197 frames at 16 fps) in length. This composite strategy allows us to generate videos retaining the appearance, occlusion, and motion realism (including camera motion) of naturally captured footage while providing the necessary ground-truth supervision for model training. In this section, we describe details on the construction of this dataset. 

\subsection{Background Clip Extraction}
\label{subsec:bgclip}
We first curated a set of full-length walking tour videos from YouTube featuring potentially empty or tranquil environments to serve as a source of background clips. To do so, we used search keywords implying emptiness, such as ``\textit{early morning},'' ``\textit{deserted downtown},'' ``\textit{lockdown street},'' and ``\textit{empty street}.'' We standardized all downloaded videos to a resolution of 720$\times$1280 and a maximum frame rate of 30~fps. To ensure structural diversity for accurate background generation, we sourced 64 background videos (57 training, and 7 testing) from 50 cities around the world (see Fig.~\ref{fig:location-map}, covering major cities, college towns, and smaller cities with different architectural styles. 

We next extracted non-overlapping $7$-second background clips from these videos. An ideal background clip will depict no humans or other dynamic objects (besides the camera wearer), such as moving cars or birds. However, this condition is too stringent to source a sufficient number of clips, and is also not easy to quantify because it requires accurately separating object motions from camera motion. We therefore relaxed our filtering process to only consider the number of humans in each video, quantified using Grounded-SAM-2~\cite{ren2024grounded} with a ``person'' text prompt. In particular, we set a threshold $P$ on the maximum number of people allowed in a single frame, and only select videos in which the percentage of frames with a human count exceeding this threshold is less than some tolerance $\tau$. We use a soft tolerance to account for noisy errors of the automatic detectors, and to permit videos with perceptually small humans at a distance from the camera wearer. We set $P=5$, and $\tau=10\%$.

We applied final quality control checks to remove poor clips with artifacts and other undesirable properties. We first used luminance-based filtering and scene-change detection heuristics to eliminate inconsistent or low-quality clips. For luminance, we compute the average Y channel value over all frames, $\bar{Y}$, and keep only clips where $\bar{Y}\in[50, 200]$. For scene changes, we compute $\operatorname{SSIM}(I_t,I_{t+1})$ between adjacent frames to capture structural differences, and the correlation $\rho(H_t,H_{t+1})$ between normalized grayscale histograms to capture global appearance shifts. A scene transition is detected if $\operatorname{SSIM}(I_t,I_{t+1}) < 0.3$ or $\rho(H_t,H_{t+1}) < 0.5$. Next, we used an Multimodal LLM~\cite{yao2024minicpm} to identify and filter out clips containing subtitles, animated stickers, and overlaid text. Finally, we conducted a manual review of the automatically filtered set, selecting 1,000 training and 35 testing clips that exhibit minimal overlap, stable viewpoints, and non-abrupt camera motions. The training and testing clips come from completely different initial raw videos. 

\subsection{Foreground Clip Extraction} 
\label{subsec:fgclip}
We sourced 10 videos from 10 cities with walking humans while qualitatively ensuring sufficient variation in the number of pixels covered by humans, from few individuals to crowds. We next extracted non-overlapping $7$-second clips from the full-length videos. Ideal foreground clips will contain one or more humans with possible associated objects (e.g., bag), and a set of foreground clips should provide sufficient diversity and duration in terms of fraction of human occupation. 
We extracted valid foreground clips by setting a lower threshold $M$ on the minimum number of frames that must have at least one detected mask using Grounded-SAM-2~\cite{ren2024grounded} with a ``person, bag, backpack'' text prompt. We set $M=138$, corresponding to $70\%$ of the clip duration. We then measured each clip's \emph{Crowd\%}, or the \emph{the mean mask area over all its frames}, and assigned each clip to one of five ranges:  0-10\%, 10-20\%, 20-30\%, 30-40\%, 40-50\%. We randomly sampled 200 clips from each range to yield 1,000 foreground clips with an even distribution over crowd sizes. 

For evaluation, we curated a set of 7 test videos from 7 cities to serve as background videos, from which we extracted foreground clips covering all \emph{Crowd\%} ranges. We generated composite clips and manually selected one clip per \emph{Crowd\%} range for each location, resulting in 35 clips in total. We selected clips with realistic depth and spatial alignment, filtering out cases with floating or disappearing humans and their associated objects.

\subsection{Composite Scene Generation}

\textbf{Shadow Simulation}
Simply cropping and pasting segmented human masks onto a background fails to account for associated visual effects such as shadows. To enhance realism, we simulated shadows for each segmented person, as illustrated in Figure~\ref{fig:data_pipeline}. For each instance, we first estimated a pivot point that defines the physical contact between the object and the ground. Using this pivot, we generated a shadow geometry by applying a combination of horizontal flipping (no flip for angles $<90^\circ$, and left–right flip for angles $\geq90^\circ$) followed by rotation around the pivot. In addition to random rotation, we applied random horizontal shear ($s_x \sim \mathcal{U}(0.15, 0.35)$) and vertical scaling ($s_y \sim \mathcal{U}(0.8, 0.95)$) to simulate variations in light direction. We sampled a single random shadow direction per clip to maintain consistent lighting across all frames and instances. We then convolved the resulting binary shadow map with a Gaussian kernel to produce a soft shadow commonly observed in the real world.\\

\noindent\textbf{Compositing and Training Tuple Construction}
To synthesize a final composited scene, we first darkened the background using the generated shadow map with a randomly sampled shadow strength $\alpha \in [0.2, 0.8]$, as shown in Figure~\ref{fig:shadow-alpha}. We then overlaid the segmented human foreground onto the shadowed background with full opacity ($\alpha = 1$). We paired each generated composite clip with its corresponding clean background clip and the original foreground mask to form a triplet: (input, mask, ground truth). The foreground masks exclude the shadow regions, allowing the model to implicitly learn the association between humans and their cast shadows during training.
\section{The \name{} Model}
\label{sec:methods}
\begin{figure*}[t!]
    \centering
    \includegraphics[width=\textwidth]{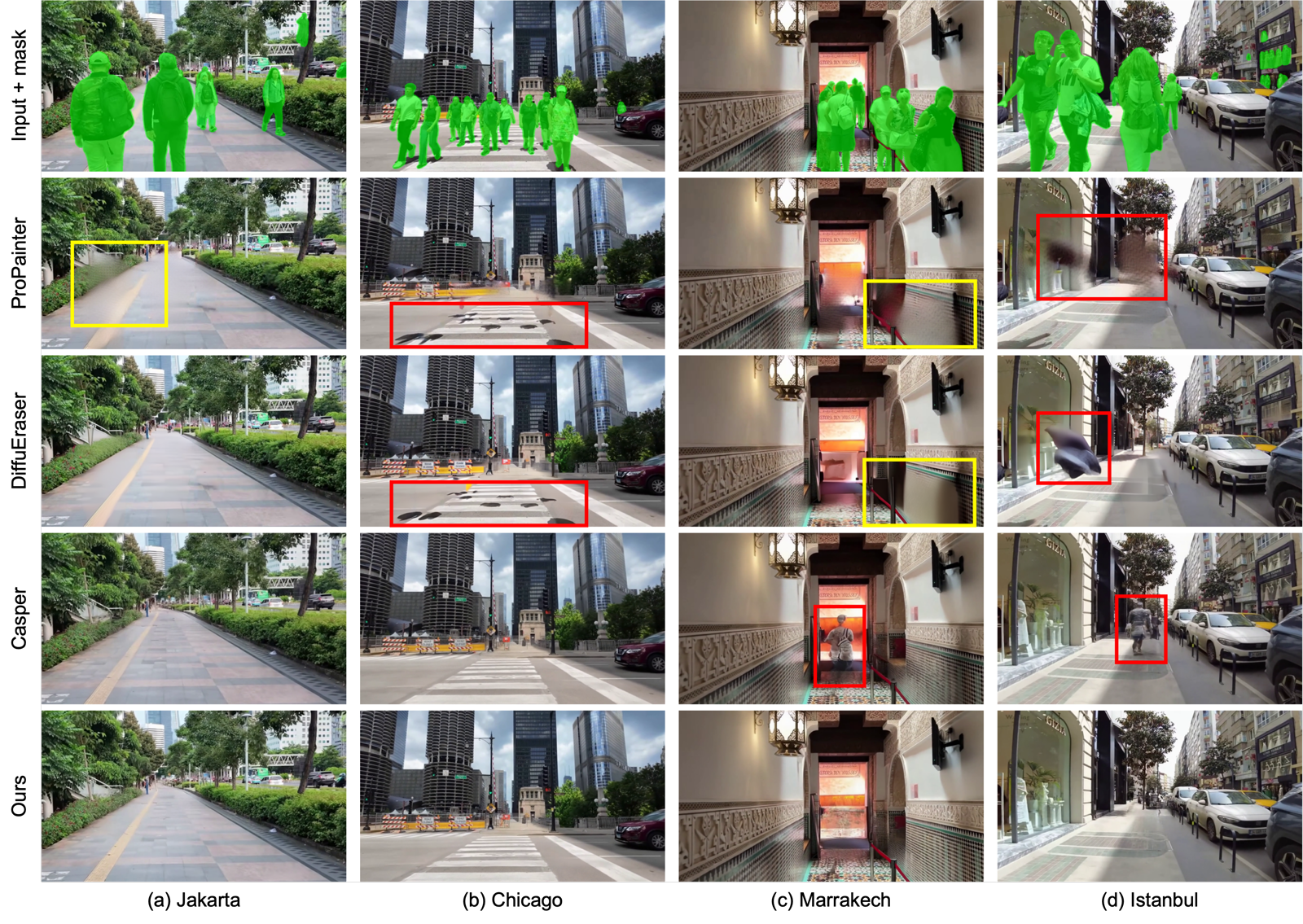}
    \caption{\textbf{Qualitative comparison.} Red boxes indicate failures to remove humans or their shadows and yellow boxes highlight areas where the background is over-smoothed instead of being filled with plausible content. When there are fewer people and the background is relatively simple, as in (a) Jakarta, all methods perform reasonably well. However, performance degrades as masks become larger and backgrounds more complex. Also, in particular, ProPainter~\cite{zhou2023propainter} and DiffuEraser~\cite{li2025diffueraser} struggle when the cast shadow is sharp (i.e., less diffused). Casper~\cite{Lee_2025_CVPR} is robust at associating effects, but when the mask is large it often hallucinates objects or people inside the masked region. In contrast, our method shows greater robustness for large masks, preserving background structure with fewer noticeable artifacts.}
    \label{fig:visual_results}
\end{figure*}
\begin{table*}[t]
\centering
\caption{
    \textbf{Quantitative comparison across varying \emph{crowd\%} levels.}
    We evaluate the removal-inpaint quality on our synthetic dataset of 7 different cities (35 videos in total). Higher PSNR indicates better reconstruction quality, while lower LPIPS and DreamSim indicate higher perceptual similarity. Best results are highlighted in red and second-best in yellow.
    }
\setlength{\tabcolsep}{3.5pt}
\renewcommand{\arraystretch}{1.15}
\resizebox{\textwidth}{!}{
    \begin{tabular}{lcccccccccccccccccc}
    \toprule
    \multirow{2}{*}{\centering\textbf{Crowd\%}} &
    \multicolumn{3}{c}{(0--10\%)} &
    \multicolumn{3}{c}{(10--20\%)} &
    \multicolumn{3}{c}{(20--30\%)} &
    \multicolumn{3}{c}{(30--40\%)} &
    \multicolumn{3}{c}{(40--50\%)} &
    \multicolumn{3}{c}{Average} \\
    & PSNR$\uparrow$ & LPIPS$\downarrow$ & DreamSim$\downarrow$
    & PSNR$\uparrow$ & LPIPS$\downarrow$ & DreamSim$\downarrow$
    & PSNR$\uparrow$ & LPIPS$\downarrow$ & DreamSim$\downarrow$
    & PSNR$\uparrow$ & LPIPS$\downarrow$ & DreamSim$\downarrow$
    & PSNR$\uparrow$ & LPIPS$\downarrow$ & DreamSim$\downarrow$
    & PSNR$\uparrow$ & LPIPS$\downarrow$ & DreamSim$\downarrow$ \\
    \midrule
    ProPainter~\cite{zhou2023propainter} & 31.81 & \second{0.058} & 0.011 & \second{28.87} & 0.088 & 0.019 & \second{25.50} & 0.128 & 0.056 & 21.98 & 0.192 & 0.114 & 20.93 & 0.238 & 0.134 & 25.82 & 0.141 & 0.067 \\
    DiffuEraser~\cite{li2025diffueraser} & \second{31.91} & \best{0.055} & \second{0.009} & 28.73 & \best{0.079} & \second{0.014} & 24.68 & \second{0.112} & 0.035 & \second{22.18} & \second{0.171} & 0.067 & \second{22.27} & \second{0.181} & 0.062 & \second{25.95} & \second{0.120} & 0.037 \\
    Casper~\cite{Lee_2025_CVPR} & 31.41 & 0.073 & 0.010 & 28.72 & 0.092 & \second{0.014} & 23.97 & 0.118 & \second{0.025} & 19.88 & 0.185 & \second{0.052} & 19.24 & 0.185 & \second{0.052} & 24.64 & 0.130 & \second{0.029} \\
    \textbf{Ours} & \best{32.39} & 0.065 & \best{0.007} & \best{29.36} & \second{0.086} & \best{0.011} & \best{26.31} & \best{0.108} & \best{0.020} & \best{22.34} & \best{0.170} & \best{0.039} & \best{23.31} & \best{0.161} & \best{0.031} & \best{26.74} & \best{0.118} & \best{0.022} \\
    \bottomrule
    \end{tabular}
    }
\label{tab:per-crowd}
\end{table*}

Given an input egocentric video clip 
and its corresponding clip of human masks 
our goal is to generate a clean, spatially and temporally consistent background video clip 
with dynamic human elements and their associated effects removed. We finetune the Casper~\cite{Lee_2025_CVPR} video diffusion model for object and effect removal in videos on the crowded and empty scene pairs in \dsname{} to perform this task. As demonstrated in~\cite{Lee_2025_CVPR}, Casper is inherently able to capture object-effect associations through appropriate large-scale pretraining. Given an input video and corresponding object masks for each instance, Casper generates a clean background and a set of single-object videos for video layering. For our task, we only require the generated backgrounds. 

We used a composite loss function to finetune Casper. The base diffusion denoising loss function is:
\begin{equation}
\mathcal{L}_{\text{base}} = \| \hat{\epsilon}_t - \epsilon_t \|_2^2,
\end{equation}
where $\hat{\epsilon}_t$ is model-predicted noise and $\epsilon_t$ is the ground-truth noise at frame $t$. To further encourage smooth temporal dynamics, we used an additional motion loss function that constrains the temporal differences of the predicted noise residuals for consecutive frames:
\begin{equation}
\mathcal{L}_{\text{sub}} = \| (\hat{\epsilon}_{t+1} - \hat{\epsilon}_t) - (\epsilon_{t+1} - \epsilon_t) \|_2^2,
\end{equation}
which penalizes discrepancies between the the temporal derivative of the noise along the frame axis. Our combined loss function is:
\begin{equation}
\mathcal{L} = (1 - \alpha)\,\mathcal{L}_{\text{base}} + \alpha\,\mathcal{L}_{\text{sub}},
\end{equation}
where $\alpha$ is the motion sub loss ratio (we set $\alpha = 0.25$). 

At inference time, the inputs to Casper (and therefore, \name{}) are an input video clip, a corresponding video clip with humans masked out, and a text prompt describing the target inpainted scene (we use \textit{``A video of a beautiful empty, human-free scene.''} for all our experiments). 

\section{Experiments}
\label{sec:experiments}

\begin{figure*}[t!]
    \centering
    \includegraphics[width=\linewidth]{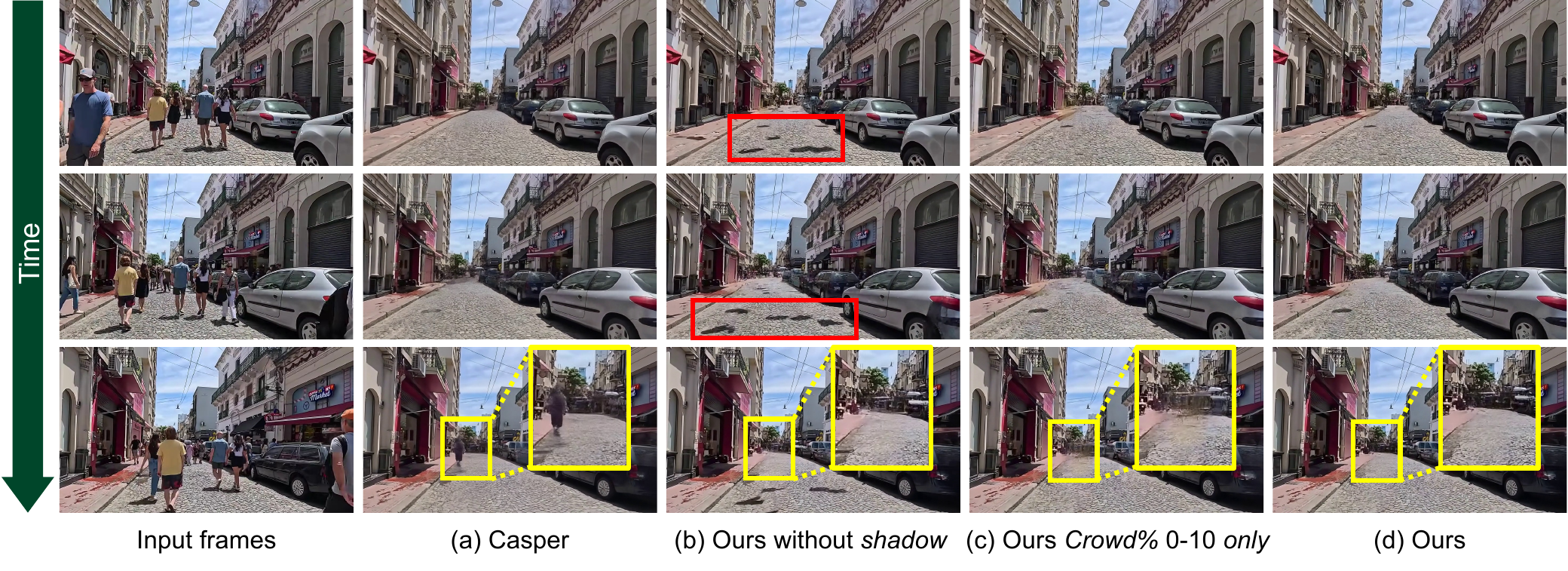}
    \caption{\textbf{Ablation results across temporal frames.} Red boxes indicate failures in capturing shadows, and the yellow boxes highlight a zoomed-in region for inpainting comparison. (a) Vanilla Casper output (baseline). (b) Finetuned on our data without shadow simulation. (c) Trained using only smaller masks (\emph{Crowd\%} 0–10). (d) Our full finetuned model with shadow injection and uniform mask distribution across different \emph{Crowd\%} levels.}
    \label{fig:ablation}
\end{figure*}
\begin{table}[t]
    \centering
    \caption{
    \textbf{Ablation study on model components} during the data construction. The first row shows the original Casper model without finetuning. Without `Shadow' drops all shadow injection, and without `Full \emph{Crowd\%}' only uses masks with lower occupancy levels (\emph{Crowd\% 0–10}). \vspace{-5pt}
    }
    \renewcommand{\arraystretch}{1.1}
    \setlength{\tabcolsep}{8pt}
    \resizebox{\columnwidth}{!}{
    \begin{tabular}{cccccc}
    \toprule
    \multicolumn{2}{c}{\textbf{Component Enabled}} 
    & \multirow{2}{*}{\textbf{PSNR}$\uparrow$} 
    & \multirow{2}{*}{\textbf{SSIM}$\uparrow$} 
    & \multirow{2}{*}{\textbf{LPIPS}$\downarrow$} 
    & \multirow{2}{*}{\textbf{DreamSim}$\downarrow$} \\
    Shadow & Full \emph{Crowd\%} & & & & \\
    \midrule
    \xmark & \xmark &  24.64  &  0.868  &  0.130  &  0.029  \\
    \xmark & \checkmark &  25.09  &  0.870  &  0.128  &  0.028  \\
    \checkmark & \xmark &  26.60  &  0.882  &  0.120  &  0.024  \\
    \checkmark & \checkmark &  26.74  &  0.881  &  0.118  &  0.022  \\
    \bottomrule
    \end{tabular}
    }
    \label{tab:ablation}
\end{table}

We used the publicly available implementation of Casper~\cite{Lee_2025_CVPR} based on the CogVideoX~\cite{yang2024cogvideox} diffusion model. We loaded the Casper model, froze its VAE and text encoder, and finetuned layers of its 3D transformer for 100 epochs on \dsname{}. Training took roughly 15 hours using 4 H200 GPUs. We compared \name{} to three baseline models: Casper, ProPainter~\cite{zhou2023propainter}, and DiffuEraser~\cite{li2025diffueraser}.  ProPainter is a video inpainting framework that combines flow-based propagation with a spatiotemporal Transformer~\cite{zhang2021spatiotemporal}. DiffuEraser is a video inpainting method based on Stable Diffusion~\cite{podell2023sdxl}, and also utilizes ProPainter as a prior.

\subsection{Results}
We first quantitatively evaluated the environment reconstruction quality of \name{} using the \dsname{} test set, which spans 7 cities and 5 \emph{Crowd\%} ranges. We selected one video per \emph{Crowd\%} range for each city. During inference, we set each video to a resolution of 720$\times$1080. We report PSNR, LPIPS, and DreamSim metrics in Table~\ref{tab:per-crowd} for all methods. 
\name{} consistently achieves the best performance on PSNR and DreamSim of all methods across all \emph{Crowd\%} ranges, while baseline methods are only comparable at low \emph{Crowd\%} settings.

We demonstrate corresponding qualitative results for generated scenes from four cities in Fig.~\ref{fig:visual_results}. When the scene contains fewer people and the background is relatively simple, as in ``Jakarta,'' all methods perform reasonably well. However, performance degrades as masks grow larger and backgrounds become more complex. ProPainter and DiffuEraser struggle with sharp cast shadows and often blur or lose detail in large masked regions. Casper handles effect associations better, but larger masks cause significant hallucinations within the masked area. In contrast, \name{} remains robust with large masks, preserving background structure and producing more visually plausible results. We provide further visual examples against baselines in Supplementary.

Next, we performed an ablation study to evaluate the individual contributions of algorithmic decisions to \name{} performance in Table~\ref{tab:ablation}. When we train the model on the dataset without shadow injections, it performs well at filling in large masks, but loses the ability to correctly associate shadows. When we train the model on smaller masks, it fails for video clips in which a crowd walks ahead of the camera wearer for all frames, leaving parts of the background constantly occluded. For instance, in Fig.~\ref{fig:ablation}, the main subject occludes the center of the frame for most of the clip's duration, resulting in color bleeding from the person wearing the yellow shirt. The uniformly distributed foreground data across \emph{Crowd\%} ranges strengthens the model’s ability to inpaint structural objects, preventing the blurring or darkening effects that commonly occur at high \emph{Crowd\%}. 

\subsection{3D Environment Modeling}
We finally qualitatively demonstrate 3D environment model extraction on several of our generated empty scenes. We reconstructed environments from both the original and crowd-removed videos using SpatialTrackerV2~\cite{xiao2025spatialtrackerv2}, a 3D point tracking model which estimates camera motion/pose, 3D scene geometry and pixel-wise 3D point trajectories given a video. We used this 4D reconstruction model to ensure a fair comparison to baselines, after finding that raw videos with human motion cause standard 3D reconstruction methods to fail. We present sample visual results in Fig.~\ref{fig:4d-results}, with additional results provided in Supplementary. When applied directly to raw walking tour videos, the reconstruction often becomes sparse and unstable due to prolonged occlusions and inconsistent visual observations. In contrast, using our cleaned videos as input leads to more coherent scene structure, improved temporal consistency, and richer background details (e.g., wall patterns). These results highlight that removing humans is a crucial step toward making egocentric walking tour videos practical for large-scale urban scene modeling.

\begin{figure*}[t!]
    \centering
    \includegraphics[width=\linewidth]{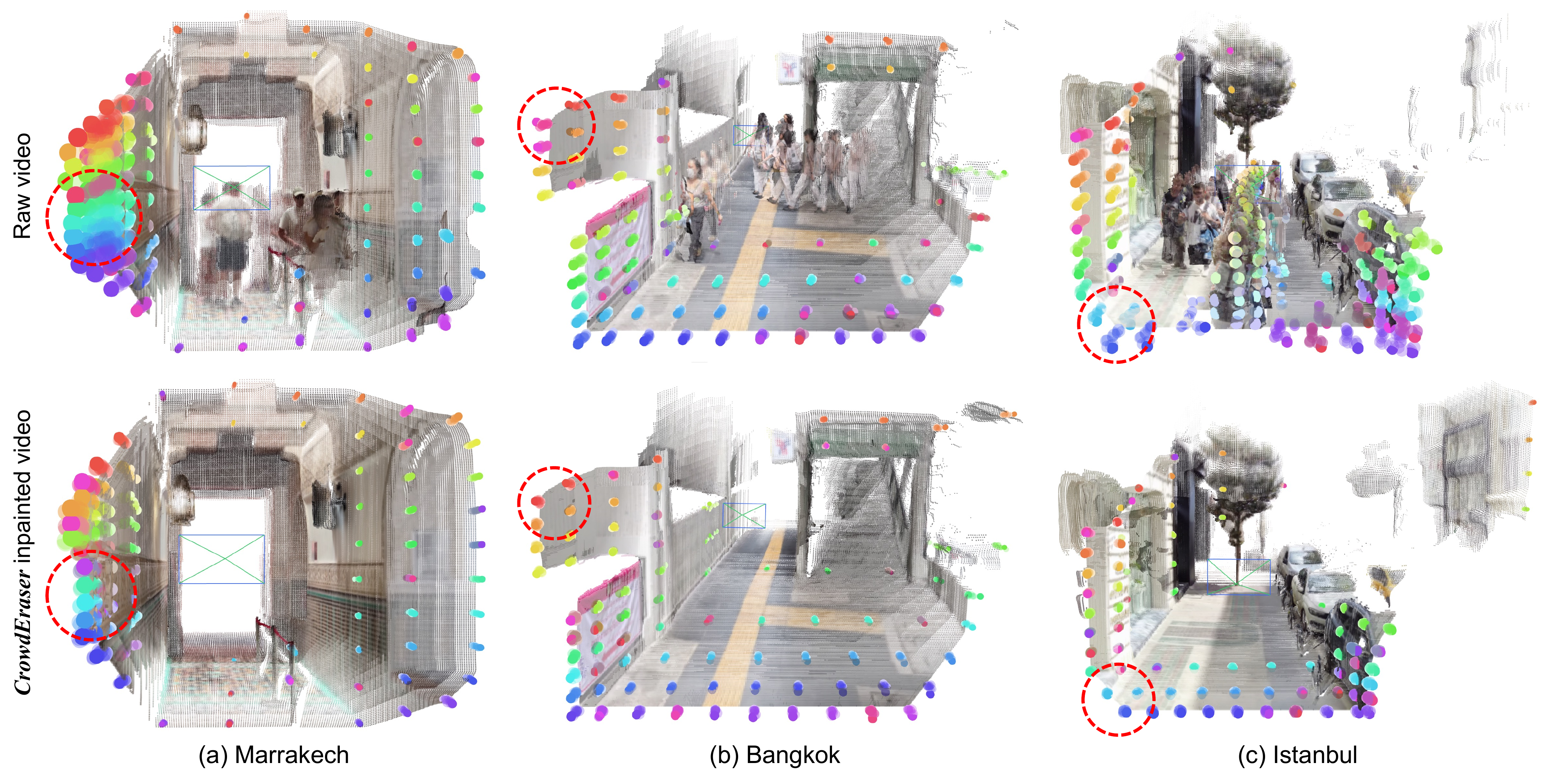}
    \caption{\textbf{SpatialTrackerV2~\cite{xiao2025spatialtrackerv2} 4D reconstruction visualization results for three scenes.} We compare results using raw walking tour video inputs (top) versus our crowd-removed versions (bottom). Each image displays the inferred 3D point clouds for the scene visualized from the camera viewpoint of the final video frame, with overlaid colored circles corresponding to point tracks in the 3D space. Points on static objects should not move. When comparing regions of static objects between the top and bottom rows (such as in the red circles), we see that point tracks exhibit greater movement, and therefore greater errors, using the raw videos. Furthermore, the \name{} cleaned videos produce better background details (e.g., wall patterns in Marrakech) and denser point clouds (e.g., right side in Istanbul). We provide further scene samples in Supplementary.}
    \label{fig:4d-results}
\end{figure*}

\begin{figure}[t!]
    \centering
    \includegraphics[width=\linewidth]{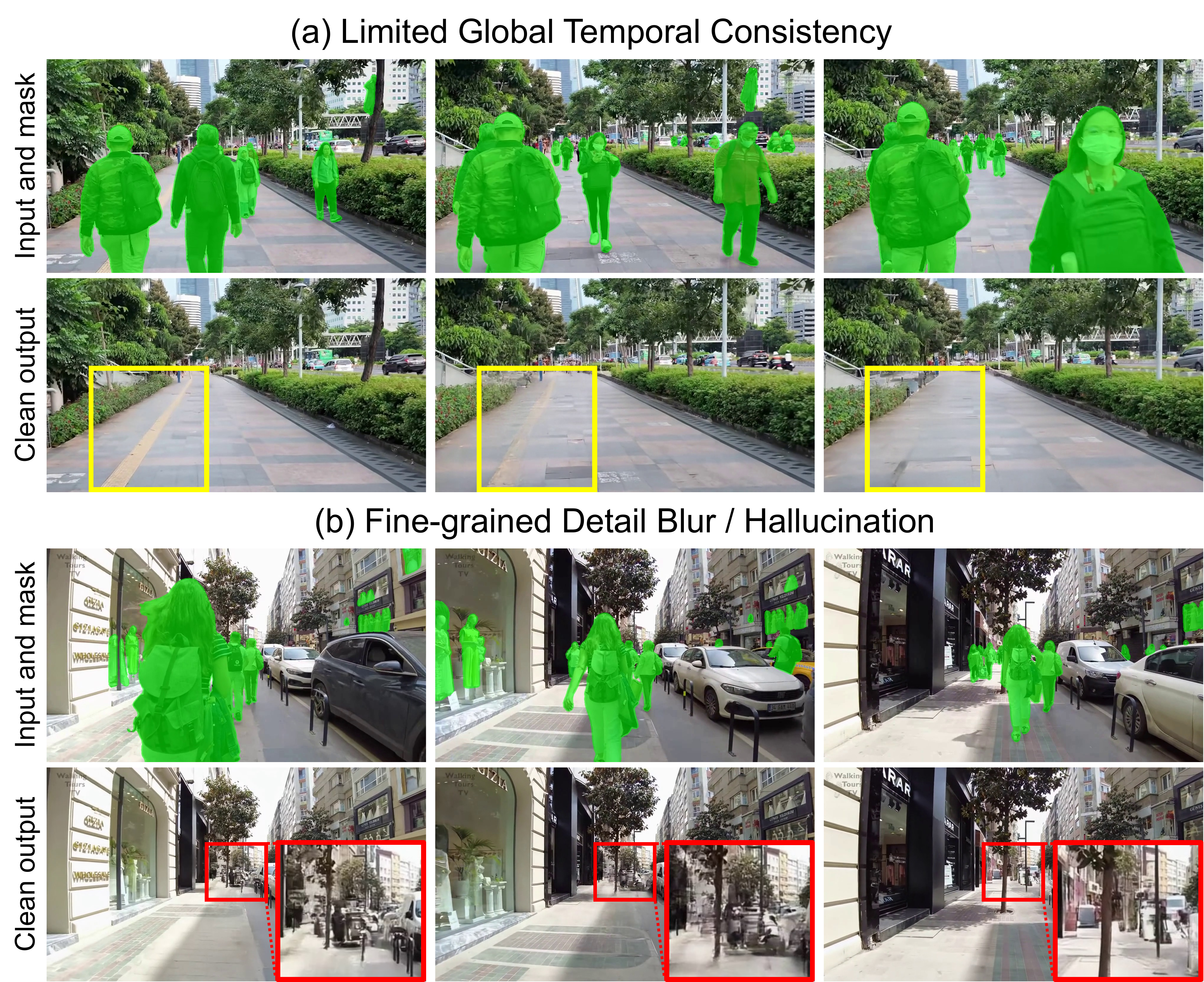}
    \caption{\textbf{Examples of difficult cases (limitations) for \name{}.} (a) For long videos with sustained large occlusions, inpainted regions lack temporal consistency across the entire video. (b) In high \emph{Crowd\%} scenarios, the model struggles to recover fine-grained, high frequency details, resulting in artifacts. \vspace{-5pt}}
    \label{fig:limitation}
\end{figure}

\section{Discussion and Conclusion}
In this work, we introduce \name{}, a diffusion-based framework designed to generate human-free environment walkthroughs from egocentric walking-tour videos. To address the lack of supervised training data for this task, we constructed \dsname{}, a semi-synthetic dataset consisting of paired crowded and empty clips sourced from real footage. This dataset enables effective fine-tuning of a video diffusion model toward large-scale crowd removal in visually diverse urban environments.

Our experiments demonstrate that \name{} significantly outperforms existing video inpainting and object-removal baselines, both quantitatively and qualitatively, particularly in challenging high \emph{Crowd\%} scenarios. Ablation studies confirm that both shadow simulation and uniform coverage across \emph{Crowd\%} levels are essential for achieving accurate effect association and robust inpainting in heavily occluded regions. Finally, we demonstrated the potential of the resulting humanless videos to enable downstream 3D urban environment modeling.

\name{} exhibits certain limitations, as shown in ~\ref{fig:limitation}. First, while urban environments present highly complex and visually rich settings, \name{} may not generalize well to scenes with visual characteristics outside of this training distribution. Second, for long videos with sustained, large occlusions, inpainted regions can lack sufficient temporal consistency. This is likely because the diffusion backbone processes fixed-length clips of 85 frames. Incorporating temporal memory mechanisms or conditioning on additional frames is a promising direction for improving long-range coherence. Finally, under high \emph{Crowd\%} scenarios, the model can struggle to reconstruct fine-grained, high-frequency details. A potential future direction is to introduce 3D geometric constraints to improve structural fidelity.


In summary, \name{} provides a practical and effective approach for generating empty environmental walkthroughs from widely available egocentric footage, expanding the potential utility of walking tour videos for various scene modeling and downstream applications.

\noindent\textbf{Acknowledgment.} Supported by the Intelligence Advanced Research Projects Activity (IARPA) via Department of Interior/ Interior Business Center (DOI/IBC) contract number 140D0423C0076. The U.S. Government is authorized to reproduce and distribute reprints for Governmental purposes notwithstanding any copyright annotation thereon. Disclaimer: The views and conclusions contained herein are those of the authors and should not be interpreted as necessarily representing the official policies or endorsements, either expressed or implied, of IARPA, DOI/IBC, or the U.S. Government.

{
    \small
    \bibliographystyle{ieeenat_fullname}
    \bibliography{main}

@String(CVPR= {IEEE Conf. Comput. Vis. Pattern Recog.})

@String(TOG= {ACM Trans. Graph.})

@String(ICLR = {Int. Conf. Learn. Represent.})

@String(AAAI = {AAAI})

@String(CVPR  = {CVPR})

@String(TOG   = {ACM TOG})

@String(ICLR  = {ICLR})

@InProceedings{Lu_2021_CVPR,
    author    = {Lu, Erika and Cole, Forrester and Dekel, Tali and Zisserman, Andrew and Freeman, William T. and Rubinstein, Michael},
    title     = {Omnimatte: Associating Objects and Their Effects in Video},
    booktitle = {Proceedings of the IEEE/CVF Conference on Computer Vision and Pattern Recognition (CVPR)},
    month     = {June},
    year      = {2021},
    pages     = {4507-4515}
}

@InProceedings{Suhail_2023_CVPR,
    author    = {Suhail, Mohammed and Lu, Erika and Li, Zhengqi and Snavely, Noah and Sigal, Leonid and Cole, Forrester},
    title     = {Omnimatte3D: Associating Objects and Their Effects in Unconstrained Monocular Video},
    booktitle = {Proceedings of the IEEE/CVF Conference on Computer Vision and Pattern Recognition (CVPR)},
    month     = {June},
    year      = {2023},
    pages     = {630-639}
}

@inproceedings{lin2023omnimatterf,
  title={Omnimatterf: Robust omnimatte with 3d background modeling},
  author={Lin, Geng and Gao, Chen and Huang, Jia-Bin and Kim, Changil and Wang, Yipeng and Zwicker, Matthias and Saraf, Ayush},
  booktitle={Proceedings of the IEEE/CVF International Conference on Computer Vision},
  pages={23471--23480},
  year={2023}
}

@article{yang2024cogvideox,
  title={Cogvideox: Text-to-video diffusion models with an expert transformer},
  author={Yang, Zhuoyi and Teng, Jiayan and Zheng, Wendi and Ding, Ming and Huang, Shiyu and Xu, Jiazheng and Yang, Yuanming and Hong, Wenyi and Zhang, Xiaohan and Feng, Guanyu and others},
  journal={arXiv preprint arXiv:2408.06072},
  year={2024}
}

@InProceedings{Lee_2025_CVPR,
    author    = {Lee, Yao-Chih and Lu, Erika and Rumbley, Sarah and Geyer, Michal and Huang, Jia-Bin and Dekel, Tali and Cole, Forrester},
    title     = {Generative Omnimatte: Learning to Decompose Video into Layers},
    booktitle = {Proceedings of the IEEE/CVF Conference on Computer Vision and Pattern Recognition (CVPR)},
    month     = {June},
    year      = {2025},
    pages     = {12522-12532}
}

@inproceedings{bar2024lumiere,
  title={Lumiere: A space-time diffusion model for video generation},
  author={Bar-Tal, Omer and Chefer, Hila and Tov, Omer and Herrmann, Charles and Paiss, Roni and Zada, Shiran and Ephrat, Ariel and Hur, Junhwa and Liu, Guanghui and Raj, Amit and others},
  booktitle={SIGGRAPH Asia 2024 Conference Papers},
  pages={1--11},
  year={2024}
}

@inproceedings{ma2024vidpanos,
  title={VidPanos: Generative panoramic videos from casual panning videos},
  author={Ma, Jingwei and Lu, Erika and Paiss, Roni and Zada, Shiran and Holynski, Aleksander and Dekel, Tali and Curless, Brian and Rubinstein, Michael and Cole, Forrester},
  booktitle={SIGGRAPH Asia 2024 Conference Papers},
  pages={1--11},
  year={2024}
}

@article{li2025diffueraser,
  title={Diffueraser: A diffusion model for video inpainting},
  author={Li, Xiaowen and Xue, Haolan and Ren, Peiran and Bo, Liefeng},
  journal={arXiv preprint arXiv:2501.10018},
  year={2025}
}

@inproceedings{lee2025video,
  title={Video diffusion models are strong video inpainter},
  author={Lee, Minhyeok and Cho, Suhwan and Shin, Chajin and Lee, Jungho and Yang, Sunghun and Lee, Sangyoun},
  booktitle={Proceedings of the AAAI Conference on Artificial Intelligence},
  volume={39},
  number={4},
  pages={4526--4533},
  year={2025}
}

@inproceedings{ding2025homogen,
  title={HomoGen: Enhanced Video Inpainting via Homography Propagation and Diffusion},
  author={Ding, Ding and Pan, Yueming and Feng, Ruoyu and Dai, Qi and Qiu, Kai and Bao, Jianmin and Luo, Chong and Chen, Zhenzhong},
  booktitle={Proceedings of the Computer Vision and Pattern Recognition Conference},
  pages={22953--22962},
  year={2025}
}

@inproceedings{zhou2023propainter,
  title={Propainter: Improving propagation and transformer for video inpainting},
  author={Zhou, Shangchen and Li, Chongyi and Chan, Kelvin CK and Loy, Chen Change},
  booktitle={Proceedings of the IEEE/CVF international conference on computer vision},
  pages={10477--10486},
  year={2023}
}

@inproceedings{bian2025videopainter,
  title={Videopainter: Any-length video inpainting and editing with plug-and-play context control},
  author={Bian, Yuxuan and Zhang, Zhaoyang and Ju, Xuan and Cao, Mingdeng and Xie, Liangbin and Shan, Ying and Xu, Qiang},
  booktitle={Proceedings of the Special Interest Group on Computer Graphics and Interactive Techniques Conference Conference Papers},
  pages={1--12},
  year={2025}
}

@inproceedings{liu2025generative,
  title={Generative video propagation},
  author={Liu, Shaoteng and Wang, Tianyu and Wang, Jui-Hsien and Liu, Qing and Zhang, Zhifei and Lee, Joon-Young and Li, Yijun and Yu, Bei and Lin, Zhe and Kim, Soo Ye and others},
  booktitle={Proceedings of the Computer Vision and Pattern Recognition Conference},
  pages={17712--17722},
  year={2025}
}

@inproceedings{rombach2022high,
  title={High-resolution image synthesis with latent diffusion models},
  author={Rombach, Robin and Blattmann, Andreas and Lorenz, Dominik and Esser, Patrick and Ommer, Bj{\"o}rn},
  booktitle={Proceedings of the IEEE/CVF conference on computer vision and pattern recognition},
  pages={10684--10695},
  year={2022}
}

@inproceedings{zhang2023adding,
  title={Adding conditional control to text-to-image diffusion models},
  author={Zhang, Lvmin and Rao, Anyi and Agrawala, Maneesh},
  booktitle={Proceedings of the IEEE/CVF international conference on computer vision},
  pages={3836--3847},
  year={2023}
}

@article{podell2023sdxl,
  title={Sdxl: Improving latent diffusion models for high-resolution image synthesis},
  author={Podell, Dustin and English, Zion and Lacey, Kyle and Blattmann, Andreas and Dockhorn, Tim and M{\"u}ller, Jonas and Penna, Joe and Rombach, Robin},
  journal={arXiv preprint arXiv:2307.01952},
  year={2023}
}

@article{li2025sekai,
  title={Sekai: A Video Dataset towards World Exploration},
  author={Li, Zhen and Li, Chuanhao and Mao, Xiaofeng and Lin, Shaoheng and Li, Ming and Zhao, Shitian and Xu, Zhaopan and Li, Xinyue and Feng, Yukang and Sun, Jianwen and others},
  journal={arXiv preprint arXiv:2506.15675},
  year={2025}
}

@inproceedings{grauman2022ego4d,
  title={Ego4d: Around the world in 3,000 hours of egocentric video},
  author={Grauman, Kristen and Westbury, Andrew and Byrne, Eugene and Chavis, Zachary and Furnari, Antonino and Girdhar, Rohit and Hamburger, Jackson and Jiang, Hao and Liu, Miao and Liu, Xingyu and others},
  booktitle={Proceedings of the IEEE/CVF conference on computer vision and pattern recognition},
  pages={18995--19012},
  year={2022}
}

@inproceedings{VenkataramananR24,
  author={Shashanka Venkataramanan and Mamshad Nayeem Rizve and João Carreira and Yuki M. Asano and Yannis Avrithis},
  title={Is ImageNet worth 1 video? Learning strong image encoders from 1 long unlabelled video},
  year={2024},
  booktitle={ICLR}
}

@article{ravi2024sam,
  title={Sam 2: Segment anything in images and videos},
  author={Ravi, Nikhila and Gabeur, Valentin and Hu, Yuan-Ting and Hu, Ronghang and Ryali, Chaitanya and Ma, Tengyu and Khedr, Haitham and R{\"a}dle, Roman and Rolland, Chloe and Gustafson, Laura and others},
  journal={arXiv preprint arXiv:2408.00714},
  year={2024}
}

@inproceedings{liu2024grounding,
  title={Grounding dino: Marrying dino with grounded pre-training for open-set object detection},
  author={Liu, Shilong and Zeng, Zhaoyang and Ren, Tianhe and Li, Feng and Zhang, Hao and Yang, Jie and Jiang, Qing and Li, Chunyuan and Yang, Jianwei and Su, Hang and others},
  booktitle={European conference on computer vision},
  pages={38--55},
  year={2024},
  organization={Springer}
}

@article{ren2024grounded,
  title={Grounded sam: Assembling open-world models for diverse visual tasks},
  author={Ren, Tianhe and Liu, Shilong and Zeng, Ailing and Lin, Jing and Li, Kunchang and Cao, He and Chen, Jiayu and Huang, Xinyu and Chen, Yukang and Yan, Feng and others},
  journal={arXiv preprint arXiv:2401.14159},
  year={2024}
}

@article{yao2024minicpm,
  title={Minicpm-v: A gpt-4v level mllm on your phone},
  author={Yao, Yuan and Yu, Tianyu and Zhang, Ao and Wang, Chongyi and Cui, Junbo and Zhu, Hongji and Cai, Tianchi and Li, Haoyu and Zhao, Weilin and He, Zhihui and others},
  journal={arXiv preprint arXiv:2408.01800},
  year={2024}
}

@inproceedings{dosovitskiy2015flownet,
  title={Flownet: Learning optical flow with convolutional networks},
  author={Dosovitskiy, Alexey and Fischer, Philipp and Ilg, Eddy and Hausser, Philip and Hazirbas, Caner and Golkov, Vladimir and Van Der Smagt, Patrick and Cremers, Daniel and Brox, Thomas},
  booktitle={Proceedings of the IEEE international conference on computer vision},
  pages={2758--2766},
  year={2015}
}

@article{wang1994representing,
  title={Representing moving images with layers},
  author={Wang, John YA and Adelson, Edward H},
  journal={IEEE transactions on image processing},
  volume={3},
  number={5},
  pages={625--638},
  year={1994},
  publisher={IEEE}
}

@inproceedings{brostow1999motion,
  title={Motion based decompositing of video},
  author={Brostow, Gabriel J and Essa, Irfan A},
  booktitle={Proceedings of the Seventh IEEE International Conference on Computer Vision},
  volume={1},
  pages={8--13},
  year={1999},
  organization={IEEE}
}

@inproceedings{fradet2008semi,
  title={Semi-automatic motion segmentation with motion layer mosaics},
  author={Fradet, Matthieu and P{\'e}rez, Patrick and Robert, Philippe},
  booktitle={European Conference on Computer Vision},
  pages={210--223},
  year={2008},
  organization={Springer}
}

@inproceedings{jojic2001learning,
  title={Learning flexible sprites in video layers},
  author={Jojic, Nebojsa and Frey, Brendan J},
  booktitle={Proceedings of the 2001 IEEE Computer Society Conference on Computer Vision and Pattern Recognition. CVPR 2001},
  volume={1},
  pages={I--I},
  year={2001},
  organization={IEEE}
}

@article{pawan2008learning,
  title={Learning layered motion segmentations of video},
  author={Pawan Kumar, M and Torr, Philip HS and Zisserman, Andrew},
  journal={International Journal of Computer Vision},
  volume={76},
  number={3},
  pages={301--319},
  year={2008},
  publisher={Springer}
}

@article{bai2009video,
  title={Video snapcut: robust video object cutout using localized classifiers},
  author={Bai, Xue and Wang, Jue and Simons, David and Sapiro, Guillermo},
  journal={ACM Transactions on Graphics (ToG)},
  volume={28},
  number={3},
  pages={1--11},
  year={2009},
  publisher={ACM New York, NY, USA}
}

@inproceedings{chuang2002video,
  title={Video matting of complex scenes},
  author={Chuang, Yung-Yu and Agarwala, Aseem and Curless, Brian and Salesin, David H and Szeliski, Richard},
  booktitle={Proceedings of the 29th annual conference on Computer graphics and interactive techniques},
  pages={243--248},
  year={2002}
}

@inproceedings{hou2019context,
  title={Context-aware image matting for simultaneous foreground and alpha estimation},
  author={Hou, Qiqi and Liu, Feng},
  booktitle={Proceedings of the IEEE/CVF International Conference on Computer Vision},
  pages={4130--4139},
  year={2019}
}

@incollection{li2005video,
  title={Video object cut and paste},
  author={Li, Yin and Sun, Jian and Shum, Heung-Yeung},
  booktitle={ACM SIGGRAPH 2005 Papers},
  pages={595--600},
  year={2005}
}

@inproceedings{sengupta2020background,
  title={Background matting: The world is your green screen},
  author={Sengupta, Soumyadip and Jayaram, Vivek and Curless, Brian and Seitz, Steven M and Kemelmacher-Shlizerman, Ira},
  booktitle={Proceedings of the IEEE/CVF Conference on Computer Vision and Pattern Recognition},
  pages={2291--2300},
  year={2020}
}

@article{wang2005interactive,
  title={Interactive video cutout},
  author={Wang, Jue and Bhat, Pravin and Colburn, R Alex and Agrawala, Maneesh and Cohen, Michael F},
  journal={ACM transactions on graphics (ToG)},
  volume={24},
  number={3},
  pages={585--594},
  year={2005},
  publisher={ACM New York, NY, USA}
}

@inproceedings{xu2017deep,
  title={Deep image matting},
  author={Xu, Ning and Price, Brian and Cohen, Scott and Huang, Thomas},
  booktitle={Proceedings of the IEEE conference on computer vision and pattern recognition},
  pages={2970--2979},
  year={2017}
}

@inproceedings{gandelsman2019double,
  title={" Double-DIP": unsupervised image decomposition via coupled deep-image-priors},
  author={Gandelsman, Yosef and Shocher, Assaf and Irani, Michal},
  booktitle={Proceedings of the IEEE/CVF conference on computer vision and pattern recognition},
  pages={11026--11035},
  year={2019}
}

@inproceedings{patwardhan2005video,
  title={Video inpainting of occluding and occluded objects},
  author={Patwardhan, Kedar A and Sapiro, Guillermo and Bertalmio, Marcelo},
  booktitle={IEEE International Conference on Image Processing 2005},
  volume={2},
  pages={II--69},
  year={2005},
  organization={IEEE}
}

@article{ebdelli2015video,
  title={Video inpainting with short-term windows: application to object removal and error concealment},
  author={Ebdelli, Mounira and Le Meur, Olivier and Guillemot, Christine},
  journal={IEEE Transactions on Image Processing},
  volume={24},
  number={10},
  pages={3034--3047},
  year={2015},
  publisher={IEEE}
}

@article{newson2014video,
  title={Video inpainting of complex scenes},
  author={Newson, Alasdair and Almansa, Andr{\'e}s and Fradet, Matthieu and Gousseau, Yann and P{\'e}rez, Patrick},
  journal={Siam journal on imaging sciences},
  volume={7},
  number={4},
  pages={1993--2019},
  year={2014},
  publisher={SIAM}
}

@article{shih2009exemplar,
  title={Exemplar-based video inpainting without ghost shadow artifacts by maintaining temporal continuity},
  author={Shih, Timothy K and Tang, Nick C and Hwang, Jenq-Neng},
  journal={IEEE transactions on circuits and systems for video technology},
  volume={19},
  number={3},
  pages={347--360},
  year={2009},
  publisher={IEEE}
}

@inproceedings{ghanbari2011contour,
  title={Contour-based video inpainting},
  author={Ghanbari, Amanna and Soryani, Mohsen},
  booktitle={2011 7th Iranian Conference on Machine Vision and Image Processing},
  pages={1--5},
  year={2011},
  organization={IEEE}
}

@article{matsushita2006full,
  title={Full-frame video stabilization with motion inpainting},
  author={Matsushita, Yasuyuki and Ofek, Eyal and Ge, Weina and Tang, Xiaoou and Shum, Heung-Yeung},
  journal={IEEE Transactions on pattern analysis and Machine Intelligence},
  volume={28},
  number={7},
  pages={1150--1163},
  year={2006},
  publisher={IEEE}
}

@inproceedings{bertalmio2001navier,
  title={Navier-stokes, fluid dynamics, and image and video inpainting},
  author={Bertalmio, Marcelo and Bertozzi, Andrea L and Sapiro, Guillermo},
  booktitle={Proceedings of the 2001 IEEE Computer Society Conference on Computer Vision and Pattern Recognition. CVPR 2001},
  volume={1},
  pages={I--I},
  year={2001},
  organization={IEEE}
}

@inproceedings{xiao2025spatialtrackerv2,
  title={Spatialtrackerv2: Advancing 3d point tracking with explicit camera motion},
  author={Xiao, Yuxi and Wang, Jianyuan and Xue, Nan and Karaev, Nikita and Makarov, Yuri and Kang, Bingyi and Zhu, Xing and Bao, Hujun and Shen, Yujun and Zhou, Xiaowei},
  booktitle={Proceedings of the IEEE/CVF International Conference on Computer Vision},
  pages={6726--6737},
  year={2025}
}

@inproceedings{li2019learning,
  title={Learning the depths of moving people by watching frozen people},
  author={Li, Zhengqi and Dekel, Tali and Cole, Forrester and Tucker, Richard and Snavely, Noah and Liu, Ce and Freeman, William T},
  booktitle={Proceedings of the IEEE/CVF conference on computer vision and pattern recognition},
  pages={4521--4530},
  year={2019}
}

@inproceedings{zhao2019data,
  title={Data augmentation using learned transformations for one-shot medical image segmentation},
  author={Zhao, Amy and Balakrishnan, Guha and Durand, Fredo and Guttag, John V and Dalca, Adrian V},
  booktitle={Proceedings of the IEEE/CVF conference on computer vision and pattern recognition},
  pages={8543--8553},
  year={2019}
}

@article{zhang2021spatiotemporal,
  title={Spatiotemporal transformer for video-based person re-identification},
  author={Zhang, Tianyu and Wei, Longhui and Xie, Lingxi and Zhuang, Zijie and Zhang, Yongfei and Li, Bo and Tian, Qi},
  journal={arXiv preprint arXiv:2103.16469},
  year={2021}
}
}

\clearpage
\setcounter{page}{1}

\twocolumn[{
\renewcommand\twocolumn[1][]{#1}%
\maketitlesupplementary

\begin{center}
    \centering \includegraphics[width=\textwidth]{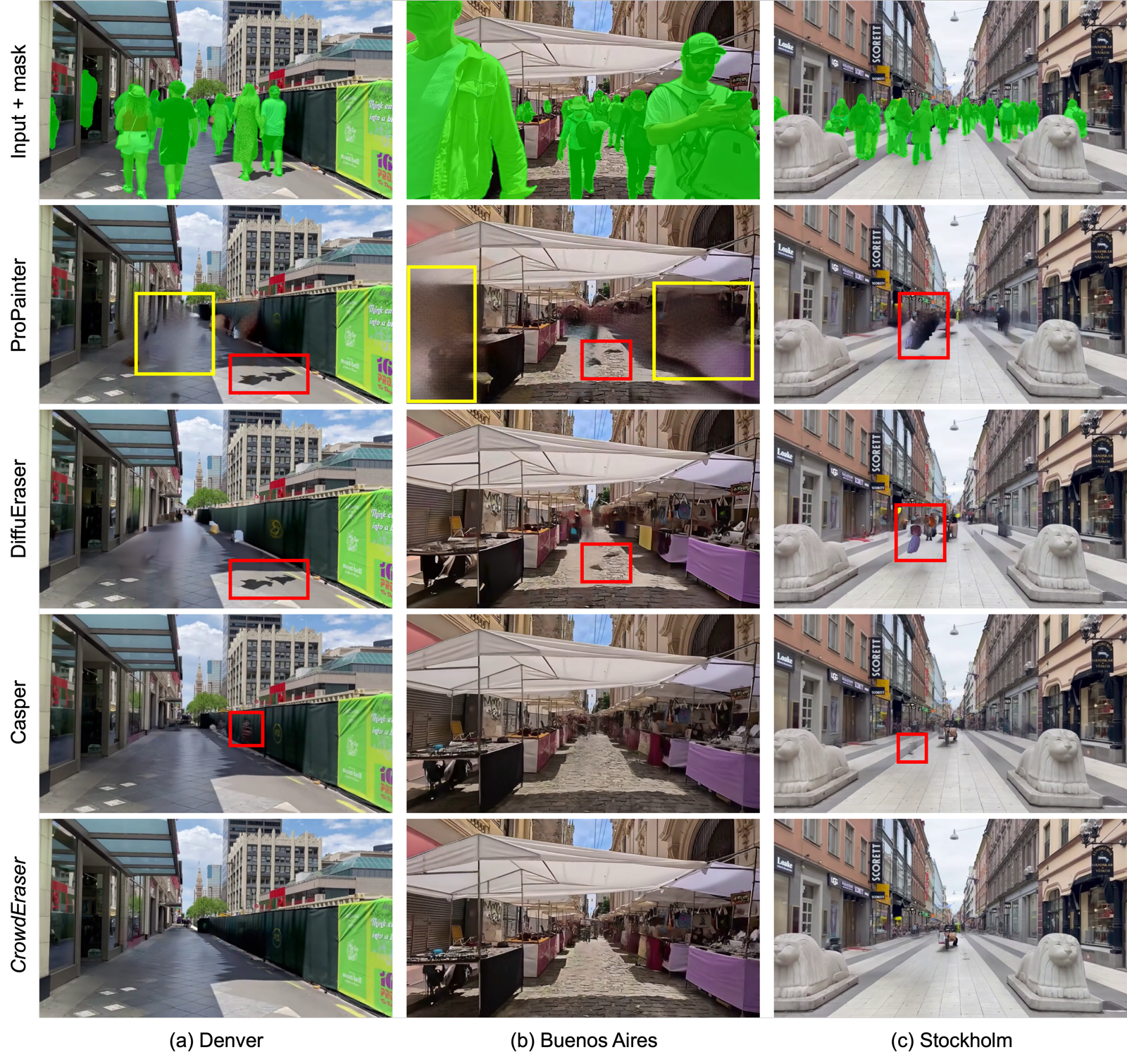}
    \captionof{figure}{\textbf{Full baseline comparison for the scenes in Figure~\ref{fig:teaser}}. Red boxes indicate failures in foreground removal or shadow handling, while yellow boxes highlight blurry inpainting artifacts. ProPainter~\cite{zhou2023propainter} and DiffuEraser~\cite{li2025diffueraser} struggle with sharp cast shadows and often blur or lose details in large masked regions. Casper achieves more reliable effect association but exhibits noticeable hallucinations when masks become large. In contrast, \name{} remains robust under significant mask sizes, preserving background structure and producing more visually plausible results.}
    \label{fig:supp-teaser}
\end{center}
}]

\twocolumn[{
\begin{center}
    \centering
    \includegraphics[width=0.97\textwidth]{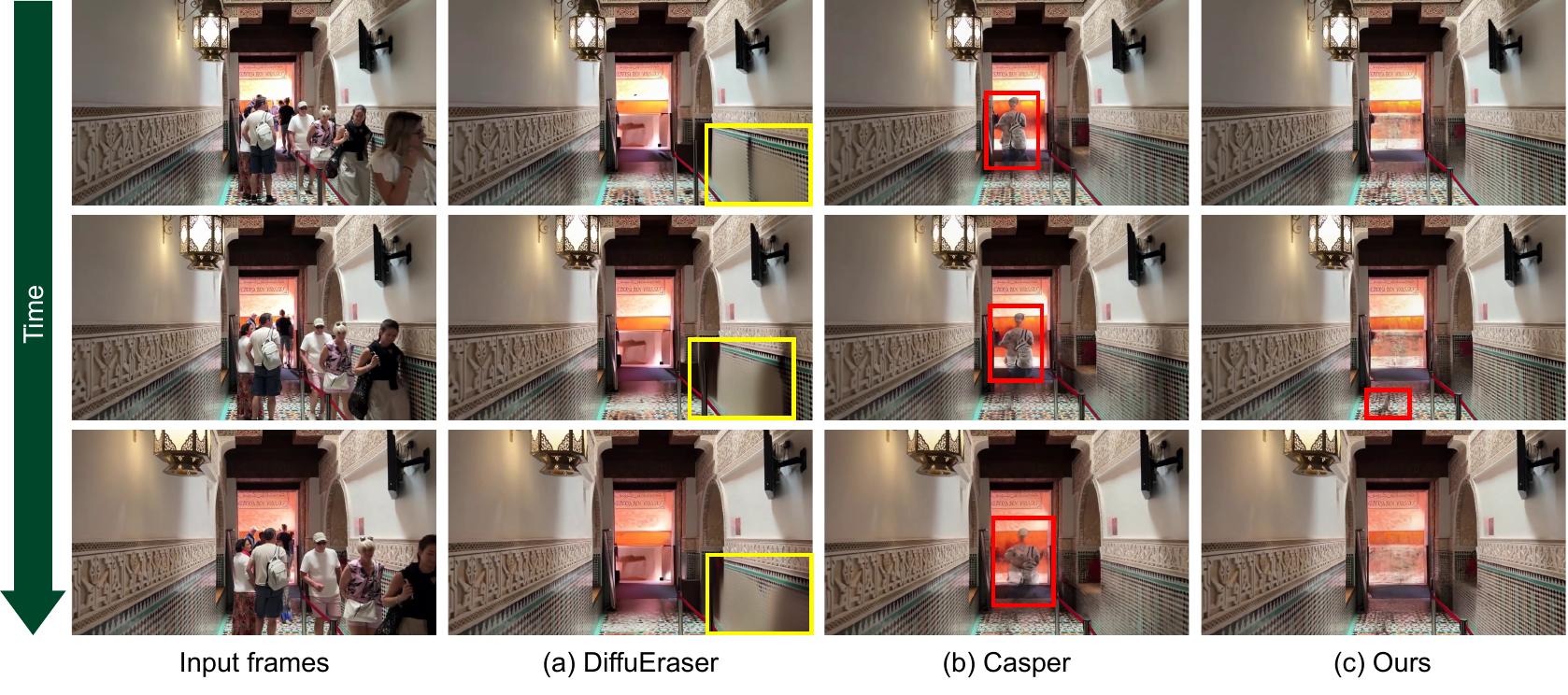}
    \captionof{figure}{\textbf{Baseline comparison across temporal frames for the ``Marrakech'' scene in Figure~\ref{fig:visual_results}}. Red boxes indicate failures in foreground removal or shadow handling, while yellow boxes mark regions where the background is over-smoothed instead of plausibly inpainted. Casper~\cite{liu2025generative} struggles with larger masks, producing noticeable hallucinations within masked areas, whereas DiffuEraser~\cite{li2025diffueraser} tends to over-smooth patterns in occluded regions.}
    \label{fig:supp-temporal3}
\end{center}

\begin{center}
    \centering
    \includegraphics[width=0.97\textwidth]{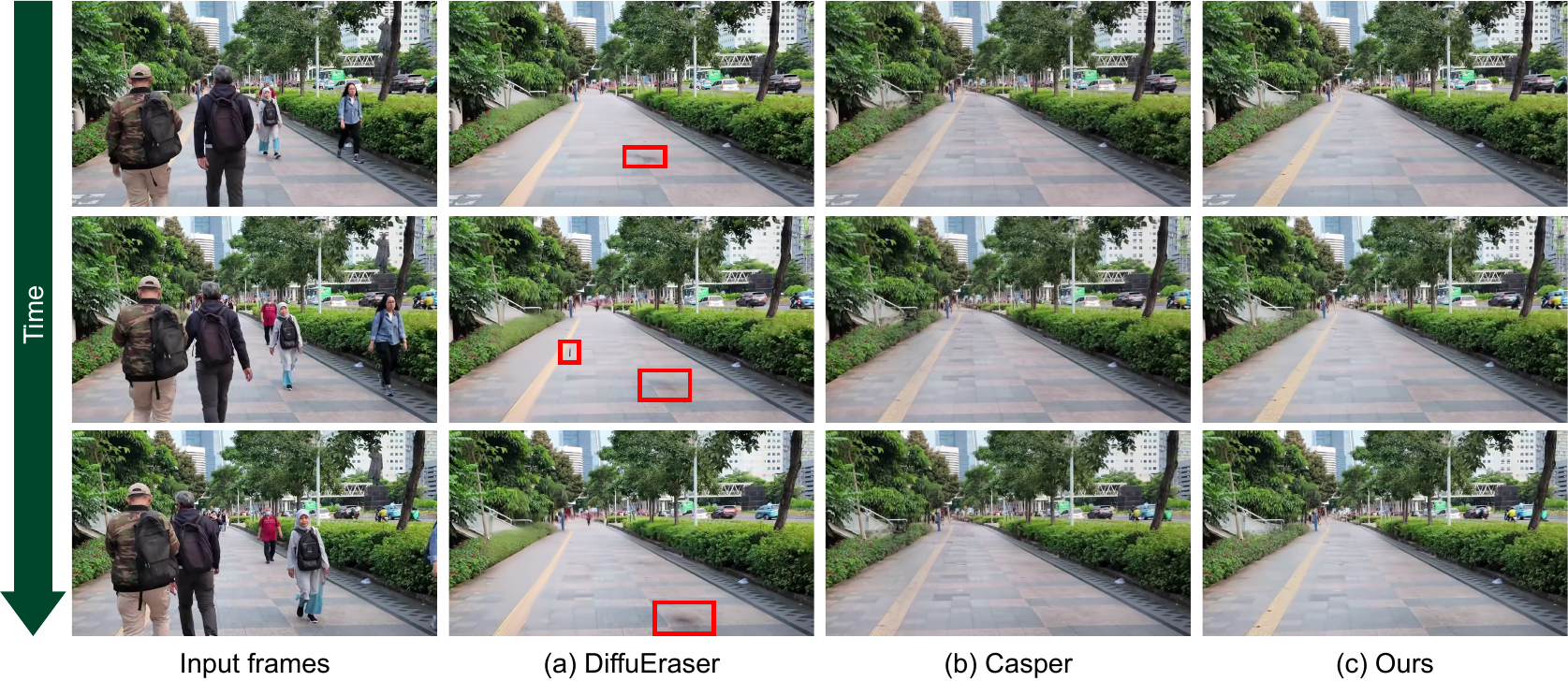}
    \captionof{figure}{\textbf{Baseline comparison across temporal frames for the `Jakarta' scene in Figure~\ref{fig:visual_results}}. Red boxes highlight failures in foreground removal or shadow handling. DiffuEraser~\cite{li2025diffueraser} struggles to capture shadows, leading to floating shadows on the pathway.}
    \label{fig:supp-temporal1}
\end{center}
}]

We provide per-scene quantitative results in Table~\ref{tab:supp-per-scene}. Our \name{} consistently achieves the best DreamSim scores, reflecting better perceptual quality aligned with human judgment. While ProPainter attains higher PSNR in some cases, this mainly stems from its strict adherence to the mask rather than improved inpainting, as PSNR is affected by unmasked, and thus unchanged, pixels. Notably, our method achieves higher in-mask PSNR, indicating more accurate inpainting within the masked regions.

\begin{figure*}[h]
    \centering
    \includegraphics[width=\textwidth]{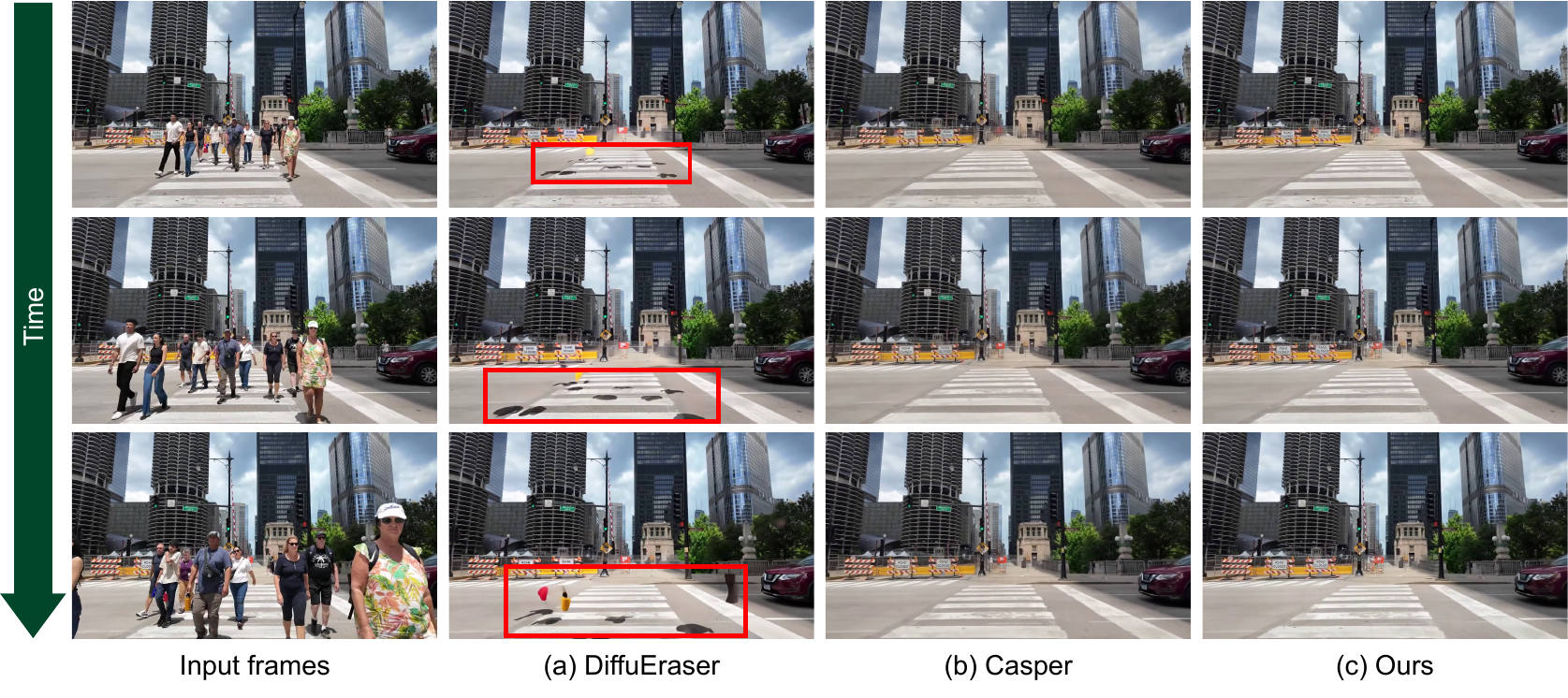}
    \caption{\textbf{Baseline comparison across temporal frames for the ``Chicago'' scene in Figure~\ref{fig:visual_results}}. Red boxes highlight failures in foreground removal or shadow handling. DiffuEraser~\cite{li2025diffueraser} struggles to associate shadows and obejects, leading to floating shadows and objects.}
    \label{fig:supp-temporal2}
\end{figure*}

\begin{figure*}[h]
    \centering
    \includegraphics[width=\textwidth]{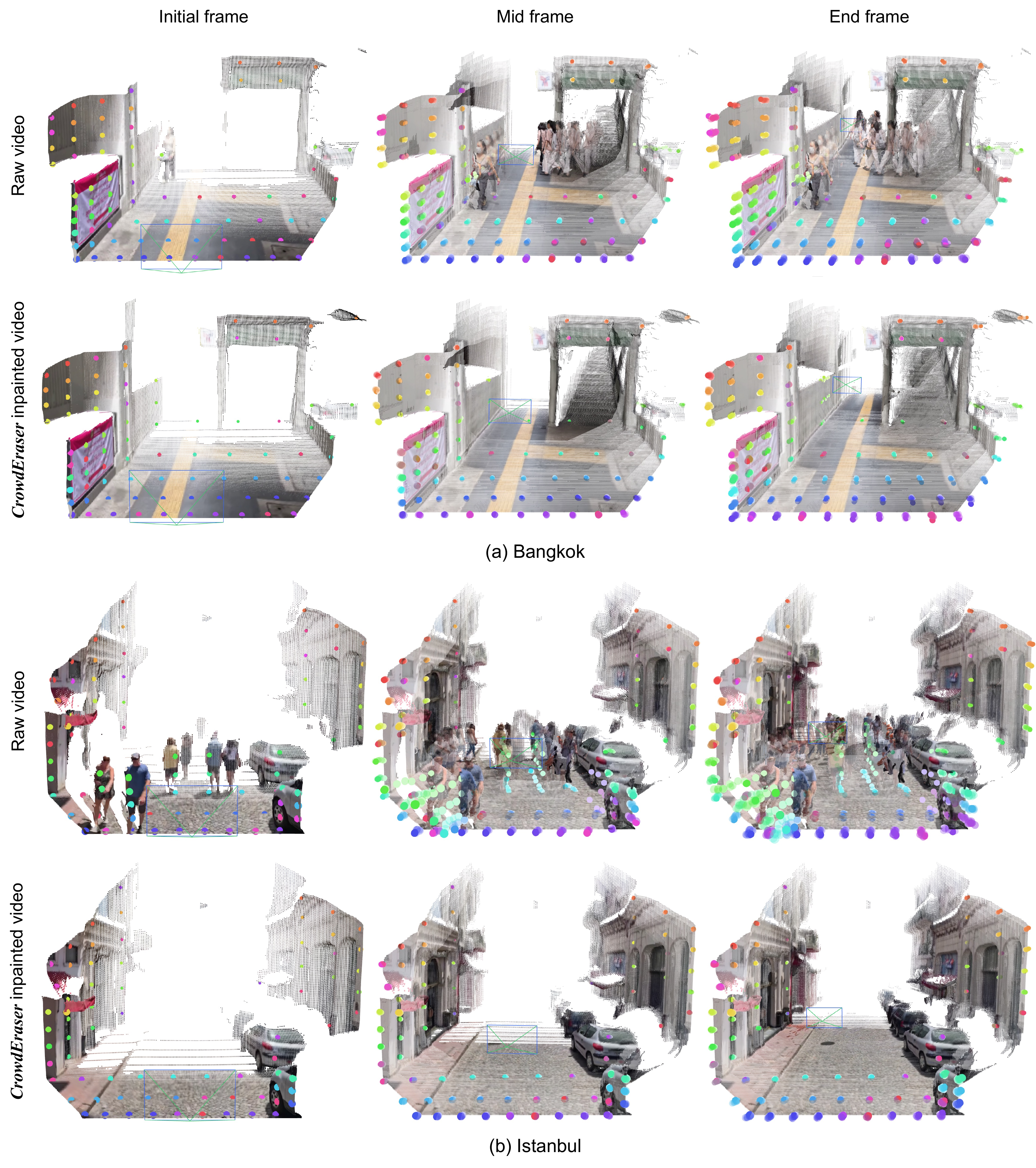}
    \caption{\textbf{SpatialTrackerV2 4D reconstruction results.} We compare results using raw walking tour video inputs (top) versus our crowd-removed versions (bottom). Each image displays the inferred 3D point clouds for the scene visualized from the camera viewpoint of the initial, middle, and final video frames, with overlaid colored circles corresponding to point tracks in the 3D space. Tracking points remain more stable in static background regions, indicating that our crowd removal leads to more reliable and robust reconstruction. Moreover, the resulting point clouds are denser and more consistent, benefiting downstream tasks such as scene modeling and 3D novel view synthesis.}
    \label{fig:supp-4drecon}
\end{figure*}

\begin{figure*}[h]
    \centering
    \includegraphics[width=\textwidth]{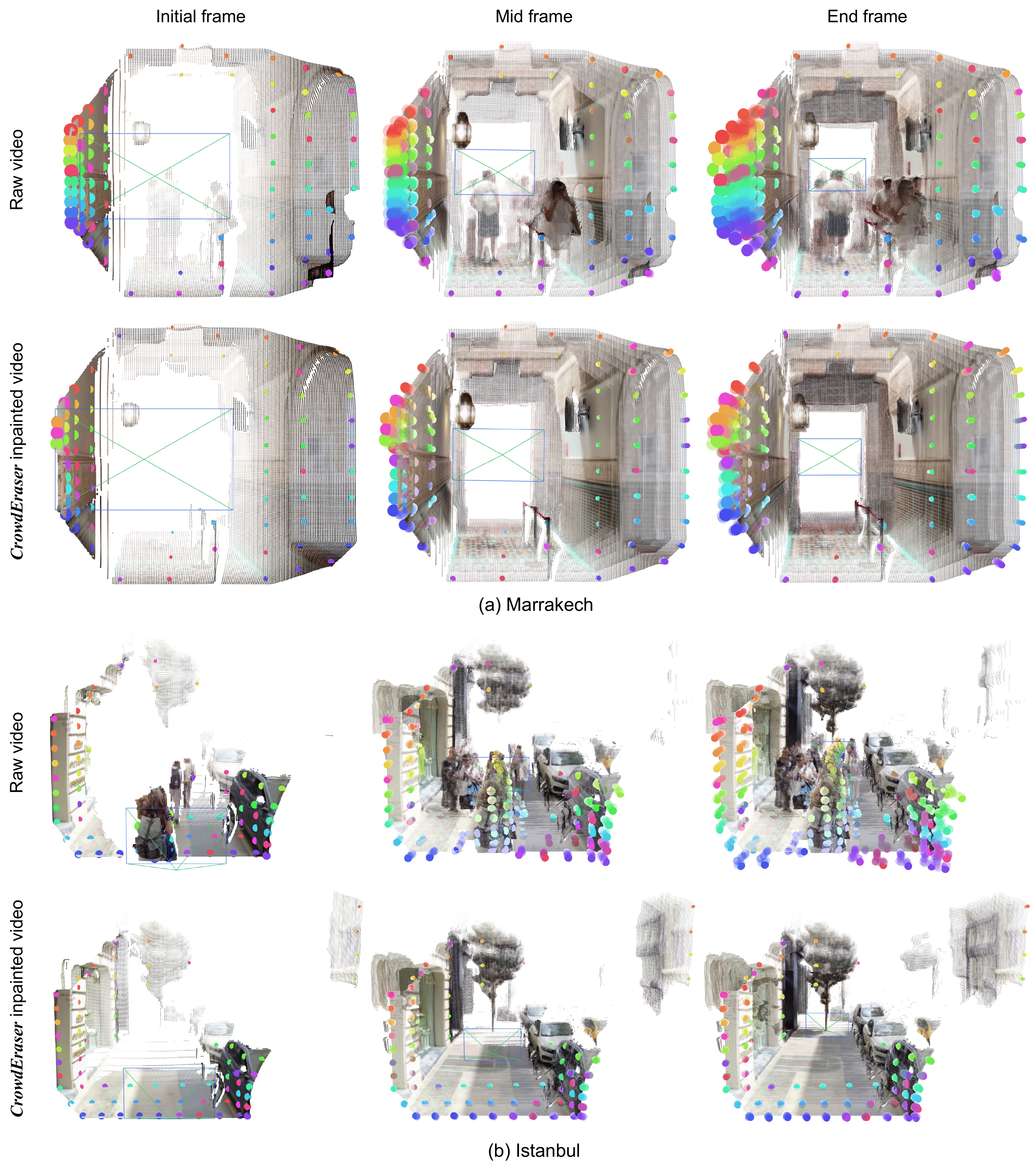}
    \caption{\textbf{SpatialTrackerV2 4D reconstruction results.} We compare results using raw walking tour video inputs (top) versus our crowd-removed versions (bottom). Each image displays the inferred 3D point clouds for the scene visualized from the camera viewpoint of the initial, middle, and final video frames, with overlaid colored circles corresponding to point tracks in the 3D space. Tracking points remain more stable in static background regions, indicating that our crowd removal leads to more reliable and robust reconstruction. Moreover, the resulting point clouds are denser and more consistent, benefiting downstream tasks such as scene modeling and 3D novel view synthesis.}
    \label{fig:supp-4drecon2}
\end{figure*}

We present additional qualitative examples, including comprehensive baseline comparisons in Figure~\ref{fig:supp-teaser}, temporal dynamics, and comparisons with two recent methods in Figures~\ref{fig:supp-temporal3}--\ref{fig:supp-temporal2}, as well as extended 4D reconstruction results in Figures~\ref{fig:supp-4drecon}--\ref{fig:supp-4drecon2}.

For reproducibility, we provide a comprehensive list of videos used to construct our dataset in Tables~\ref{tab:train-video-sources}--\ref{tab:viz-video-sources}.

\begin{table*}[h]
\centering
\caption{\textbf{Quantitative comparison across cities.} Best results are highlighted in red and second-best in yellow. Our \name{} consistently achieves the best DreamSim, reflecting superior perceptual quality.}
\setlength{\tabcolsep}{3.5pt}
\renewcommand{\arraystretch}{1.15}
\resizebox{\textwidth}{!}{%
\begin{tabular}{l
                *{7}{cc}    
                }         
\toprule
\multirow{2}{*}{\textbf{Scene}} &
\multicolumn{2}{c}{\textbf{Birmingham}} &
\multicolumn{2}{c}{\textbf{Boston}} &
\multicolumn{2}{c}{\textbf{Capetown}} &
\multicolumn{2}{c}{\textbf{Chicago}} &
\multicolumn{2}{c}{\textbf{Dubai}} &
\multicolumn{2}{c}{\textbf{Rome}} &
\multicolumn{2}{c}{\textbf{Zurich}} \\
& PSNR$\uparrow$ & DreamSim$\downarrow$
& PSNR$\uparrow$ & DreamSim$\downarrow$
& PSNR$\uparrow$ & DreamSim$\downarrow$
& PSNR$\uparrow$ & DreamSim$\downarrow$
& PSNR$\uparrow$ & DreamSim$\downarrow$
& PSNR$\uparrow$ & DreamSim$\downarrow$
& PSNR$\uparrow$ & DreamSim$\downarrow$ \\
\midrule
ProPainter~\cite{zhou2023propainter}    & \best{27.77} & 0.052 & \best{26.00} & 0.049 & \best{25.86} & 0.059 & 25.91 & 0.053 & 22.34 & 0.050 & \second{29.97} & 0.019 & \second{25.76} & 0.059 \\
DiffuEraser~\cite{li2025diffueraser}     & \second{27.25} & 0.034 & \second{25.49} & 0.042 & 25.43 & 0.050 & 25.59 & 0.036 & 22.65 & 0.042 & 29.86 & 0.013 & 25.14 & 0.049 \\
Casper~\cite{Lee_2025_CVPR}          & 26.15 & \second{0.028} & 25.12 & \second{0.031} & 24.20 & \second{0.035} & \second{26.02} & \second{0.022} & \second{23.14} & \second{0.027} & 29.44 & \second{0.012} & 25.21 & \second{0.027} \\
\textbf{Ours}           & 26.58 & \best{0.022} & \best{26.00} & \best{0.025} & \second{25.50} & \best{0.026} & \best{27.08} & \best{0.019} & \best{24.81} & \best{0.019} & \best{30.31} & \best{0.009} & \best{25.99} & \best{0.022} \\
\bottomrule
\end{tabular}%
}
\label{tab:supp-per-scene}
\end{table*}

\begin{figure*}[h]
    \centering
    \includegraphics[width=\textwidth]{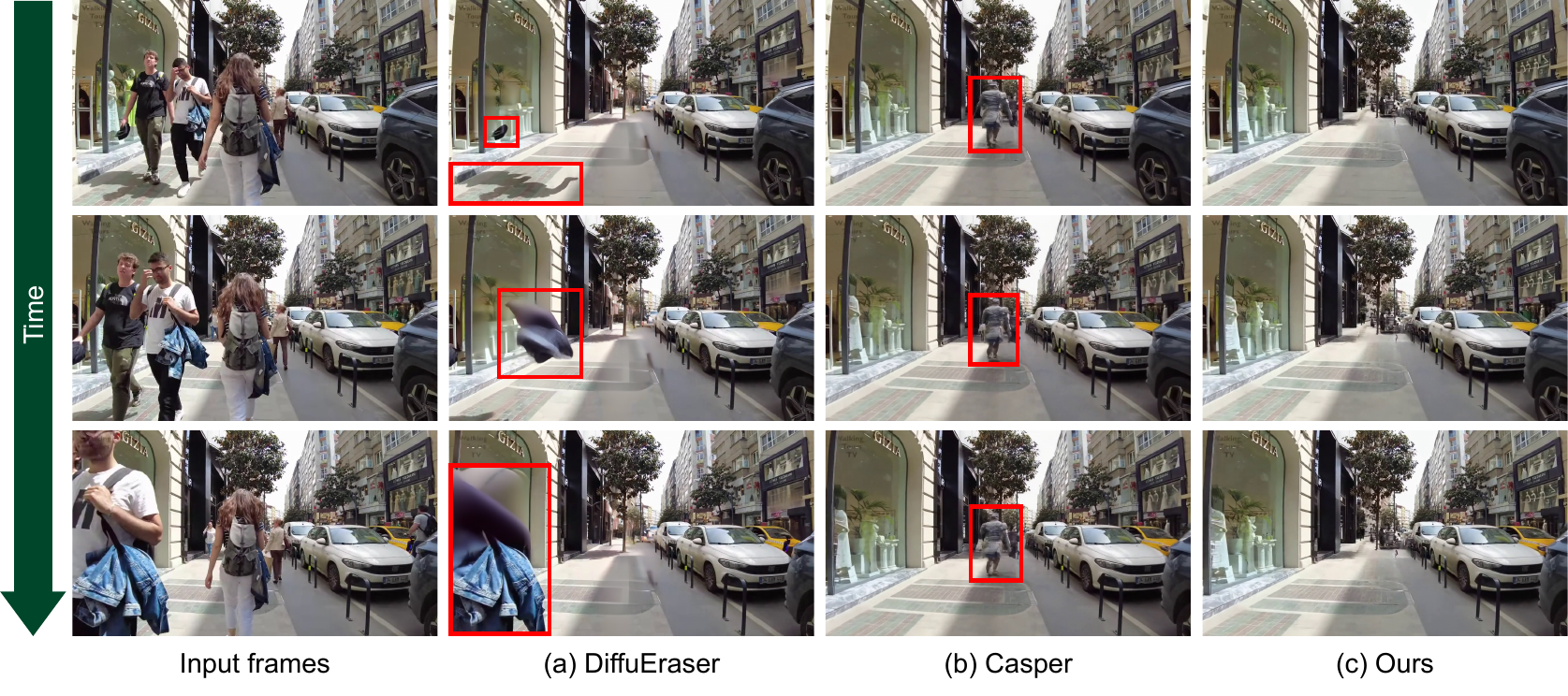}
    \caption{\textbf{Baseline comparison across temporal frames for the ``Istanbul'' scene in Figure~\ref{fig:visual_results}}. Red boxes indicate failures in foreground removal or shadow handling. Casper~\cite{liu2025generative} struggles with larger masks, producing noticeable hallucinations within masked areas, while DiffuEraser~\cite{li2025diffueraser} has difficulty handling shadows and removing associated objects.}
    \label{fig:supp-temporal4}
\end{figure*}

\begin{table*}[t]
\centering
\resizebox{\textwidth}{!}{
\begin{tabular}{llll}
\toprule
\textbf{Continent} & \textbf{Country} & \textbf{City / Area} & \textbf{URL(s)} \\
\midrule
\multicolumn{4}{l}{\textbf{Train Background Video Sources}} \\
\midrule

Africa & Egypt & Cairo & \url{https://youtu.be/TVe7Th_EfrM} \\

Asia & China & Beijing & 
\url{https://youtu.be/FNK5UEObcEg}, \url{https://youtu.be/MU-obosH1ow} \\
Asia & China & Great Wall & \url{https://youtu.be/cVmM6sUcdwg} \\
Asia & China & Shanghai &
\url{https://youtu.be/Z1i0v6wsGLI}, \url{https://youtu.be/DUANyWsER_o} \\
Asia & South Korea & Cheongju & \url{https://youtu.be/kqkUJZllt0U} \\
Asia & South Korea & Daejeon &
\url{https://youtu.be/n1V69LjdNQg}, \url{https://youtu.be/mGLh9Ss_OEo} \\
Asia & South Korea & Seoul &
\url{https://youtu.be/LJaIGjruqtE} \\

Europe & Austria & Vienna & \url{https://youtu.be/LKNyOXwooKo} \\
Europe & France & Paris & \url{https://youtu.be/HJgflqZvTl0} \\
Europe & Germany & Berlin & \url{https://youtu.be/qgNKZBQW0hA} \\
Europe & Germany & Würzburg & \url{https://youtu.be/hFzqKwrRdi8} \\
Europe & Italy & Pompeii & \url{https://youtu.be/9L1jrC2-BTE} \\
Europe & Italy & Positano & \url{https://youtu.be/UgYMsj4dDfE} \\
Europe & Spain & Majorca & \url{https://youtu.be/ufPda6XaA7E} \\
Europe & Sweden & Stockholm & \url{https://youtu.be/HJgflqZvTl0} \\
Europe & UK & London & \url{https://youtu.be/VkFpqAG6mm8} \\

North America & Canada & Guelph, ON & \url{https://youtu.be/mLk9-S6hpsU} \\
North America & Canada & North Vancouver, BC & \url{https://youtu.be/IAhJvjeM9SI} \\
North America & Canada & Surrey, BC & \url{https://youtu.be/mjr1r190-VQ} \\
North America & Canada & Toronto, ON & \url{https://youtu.be/5I-WYTYLD0o} \\

North America & USA & Atlanta, GA & \url{https://youtu.be/mseo6t1hiYs} \\
North America & USA & Austin, TX & \url{https://youtu.be/7cVQAs-c2Lg} \\
North America & USA & Boston, MA &
\url{https://youtu.be/6ZwGo49Dce8}, \url{https://youtu.be/2MHqXI-j-zY} \\
North America & USA & Cambridge, MA & \url{https://youtu.be/uHRHlba3CyQ} \\
North America & USA & Charleston, SC & \url{https://youtu.be/JdPkO2iIvfg} \\
North America & USA & Hayward, CA & \url{https://youtu.be/9EDA2IHtJFM} \\
North America & USA & Honolulu, HI & \url{https://youtu.be/JTQdSKz9wEc} \\
North America & USA & Houston, TX & \url{https://youtu.be/t2ojP7lrfXw} \\
North America & USA & Las Vegas, NV & \url{https://youtu.be/GH25Pzv0WNo} \\
North America & USA & Los Angeles, CA &
\url{https://youtu.be/kiyUR7xPkAM}, \url{https://youtu.be/EM1XQfC1Vdw} \\
North America & USA & Miami, FL & \url{https://youtu.be/ruXuOM1PAJY} \\
North America & USA & New Haven, CT & \url{https://youtu.be/r0_sbCxgP58} \\
North America & USA & New York, NY &
\url{https://youtu.be/3koOEPntvqk}, \url{https://youtu.be/2UXhhyNYpLc} \\
 & & & \url{https://youtu.be/MheS3NBAZJ0}, \url{https://youtu.be/YbiCtAdiS6U} \\
 & & & \url{https://youtu.be/fYY7uEgPw1c} \\
North America & USA & Pine Bluff, AR & \url{https://youtu.be/3FDjNp77wGo} \\
North America & USA & Portland, OR &
\url{https://youtu.be/TkZU-yfUqe8}, \url{https://youtu.be/KiZ36s2IUi0} \\
North America & USA & Provo, UT & \url{https://youtu.be/653tnKwzNdg} \\
North America & USA & Sacramento, CA & \url{https://youtu.be/W5XSfxIZdMg} \\
North America & USA & San Diego, CA & \url{https://youtu.be/m13-S2HEl6E} \\
North America & USA & San Francisco, CA & \url{https://youtu.be/SX-2d1VyTUw} \\
North America & USA & San Jose, CA &
\url{https://youtu.be/DNnP60oi-mc}, \url{https://youtu.be/Kc_NWFQrzpo} \\
North America & USA & State College, PA & \url{https://youtu.be/R81NaRZISTU} \\
North America & USA & Syracuse, NY & \url{https://youtu.be/FIg579AzSUg} \\
North America & USA & Washington, DC & \url{https://youtu.be/secTBj63dcI} \\
North America & USA & Wellesley, MA & \url{https://youtu.be/dCxAuWK5gLw} \\
North America & USA & North Dakota & \url{https://youtu.be/mr02QEJooOQ} \\

\midrule
\multicolumn{4}{l}{\textbf{Train Foreground Video Sources}} \\
\midrule

Asia & India & Varanasi & \url{https://youtu.be/Odh_7dQwzYQ} \\
Asia & South Korea & Seoul &
\url{https://youtu.be/D-F4L5Gfhik}, \url{https://youtu.be/DF8KDaUn1TA} \\
 & & & \url{https://youtu.be/KisjSKv53FA} \\

Europe & Germany & Hamburg & \url{https://youtu.be/aqgRc-sne8g} \\
Europe & Netherlands & Amsterdam & \url{https://youtu.be/7Ttc3AaPNZs} \\

North America & USA & Anaheim, CA & \url{https://youtu.be/Eo8q61Xtc50} \\
North America & USA & Honolulu, HI & \url{https://youtu.be/eSSrUot4yhQ} \\
North America & USA & New York, NY &
\url{https://youtu.be/bCoqUaLHjy0}, \url{https://youtu.be/C_nK_-ZI6Zo} \\

\bottomrule
\end{tabular}
}
\caption{\textbf{Training dataset sources.} Background and foreground video sources used to construct \dsname{}. For cities where a single video did not provide sufficient clips, multiple videos were collected to ensure sufficient coverage.}
\label{tab:train-video-sources}
\end{table*}
\begin{table*}[ht!]
    \centering
    \resizebox{0.9\textwidth}{!}{
    \begin{tabular}{llll}
    \toprule
    \textbf{Continent} & \textbf{Country} & \textbf{City / Area} & \textbf{URL(s)} \\
    \midrule
    \multicolumn{4}{l}{\textbf{Test Data Background}} \\
    \midrule
    
    Africa & South Africa & Cape Town & 
    \url{https://www.youtube.com/watch?v=eG_SV5aSBqQ} \\
    
    Asia & United Arab Emirates & Dubai &
    \url{https://www.youtube.com/watch?v=mElSLruob6c} \\
    
    Europe & Italy & Rome &
    \url{https://www.youtube.com/watch?v=xpRDEoEQpwk} \\
    Europe & Switzerland & Zurich &
    \url{https://www.youtube.com/watch?v=UcRW2OHqC2o} \\
    Europe & UK & Birmingham &
    \url{https://www.youtube.com/watch?v=IlzFH4yE2Z8} \\
    
    North America & USA & Boston, MA &
    \url{https://www.youtube.com/watch?v=8NVjJsjFLEA} \\
    North America & USA & Chicago, IL &
    \url{https://www.youtube.com/watch?v=R9VGInHbKik} \\
    
    \midrule
    \multicolumn{4}{l}{\textbf{Test Data Foreground}} \\
    \midrule
    
    Asia & China & Hong Kong &
    \url{https://www.youtube.com/watch?v=JRvQ_pM87ik} \\
    Asia & Switzerland & Zurich &
    \url{https://www.youtube.com/watch?v=65KsVRG1ao8} \\
    
    Europe & Austria & Vienna &
    \url{https://www.youtube.com/watch?v=TCRD9Dz6k88} \\
    
    North America & USA & Houston, TX &
    \url{https://www.youtube.com/watch?v=r6cLF5s2B_g} \\
    North America & USA & Los Angeles, CA &
    \url{https://www.youtube.com/watch?v=Oifxr_fLfNE} \\
    
    \bottomrule
    \end{tabular}
    }
    \caption{\textbf{Quantitative test dataset sources.} Background and foreground video sources used in the quantitative evaluation dataset.}
    \label{tab:test-video-sources}
\end{table*}

\begin{table*}[ht]
    \centering
    \resizebox{0.95\textwidth}{!}{
    \begin{tabular}{lllp{12cm}}
    \toprule
    Continent & Country & City & URL(s) \\
    \midrule
    \multicolumn{4}{l}{\textbf{Crowd Walking Tour Video Sources}} \\
    \midrule
    Asia & India & Mumbai & \url{https://youtu.be/_2GM4gVlors} \\
    Asia & Japan & Tokyo & \url{https://youtu.be/jeQd-n7Rot0} \\
    Asia & Japan & Kyoto & \url{https://youtu.be/OhOlwqjt_Lg} \\
    Asia & Thailand & Bangkok & \url{https://youtu.be/sWRoDRYi1Lk} \\
    Asia & Indonesia & Jakarta & \url{https://youtu.be/2lSUV5KZgwI} \\
    Africa & South Africa & Cape Town & \url{https://youtu.be/pL-5CjB0hf8} \\
    Africa & Morocco & Marrakech & \url{https://youtu.be/OvN1numZqqU} \\
    Africa & Nigeria & Lagos & \url{https://youtu.be/LZJ000F-CLc} \\
    North America & USA & New York City & \url{https://youtu.be/o012OW9vej8}, \url{https://youtu.be/77EXFlRLbiM} \\
    North America & USA & Denver & \url{https://youtu.be/L3Uz0O1pO3k} \\
    North America & USA & Chicago & \url{https://youtu.be/75OQ94gCOeI} \\
    North America & Mexico & Cancun & \url{https://youtu.be/3mU1CbBTIJ8}, \url{https://youtu.be/RlcMESpoHs8} \\
    South America & Brazil & Rio de Janeiro & \url{https://youtu.be/RYxqpz5XS0A} \\
    South America & Argentina & Buenos Aires & \url{https://youtu.be/Hug4u_7ZYxE}, \url{https://youtu.be/QVyueY43tA8} \\
    Europe & Croatia & Dubrovnik & \url{https://youtu.be/_93zEDEYBf0} \\
    Europe & Sweden & Stockholm & \url{https://youtu.be/FuFe9WC3rjg} \\
    Europe & Spain & Seville & \url{https://youtu.be/m4AmnRnWcRk}, \url{https://youtu.be/1o_V2qXNyUM} \\
    Europe/Asia & Turkey & Istanbul & \url{https://youtu.be/ZOGEYkHyNWU}, \url{https://youtu.be/Hp3gETuZa8o} \\
    \bottomrule
    \end{tabular}}
    \caption{\textbf{Qualitative test dataset sources.} Real walking tour video sources used for qualitative evaluation. For cities where a single video did not provide sufficient clips, multiple videos were collected to ensure sufficient coverage.}
    \label{tab:viz-video-sources}
\end{table*}


\end{document}